\documentclass[pdflatex,sn-mathphys-num]{sn-jnl}

\usepackage{graphicx}%
\usepackage{multirow}%
\usepackage{amsmath,amssymb,amsfonts}%
\usepackage{amsthm}%
\usepackage{mathrsfs}%
\usepackage[title]{appendix}%
\usepackage{xcolor}%
\usepackage{textcomp}%
\usepackage{manyfoot}%
\usepackage{booktabs}%
\usepackage{adjustbox}
\usepackage{algorithm}%
\usepackage{algorithmicx}%
\usepackage{algpseudocode}%
\usepackage{listings}%
\usepackage{xcolor}
\usepackage{subcaption}

\theoremstyle{thmstyleone}%
\newtheorem{theorem}{Theorem}
\newtheorem{proposition}[theorem]{Proposition}%

\theoremstyle{thmstyletwo}%

\theoremstyle{thmstylethree}%

\begin{document}

\title[Article Title]{ProDER: A Continual Learning Approach for Fault Classification and Localization in Evolving Smart Grids}


\author[1]{\fnm{Emad} \sur{Efatinasab}}

\author[1]{\fnm{Nahal} \sur{Azadi}}

\author[1]{\fnm{Davide} \sur{Dalle Pezze}}

\author[1]{\fnm{Gian Antonio} \sur{Susto}}

\author[2]{\fnm{Chuadhry Mujeeb} \sur{Ahmed}}

\author[1]{\fnm{Mirco} \sur{Rampazzo}}

\affil[1]{\orgdiv{Department of Information Engineering}, \orgname{University of Padova}}

\affil[2]{\orgdiv{School of Computing}, \orgname{Newcastle University}}



\abstract{
{
Data-driven fault diagnosis models for smart grids are usually trained once on a fixed dataset, whereas in operation new fault types appear and monitoring is extended to new grid zones. Retraining from scratch on all accumulated data is costly, while naively updating the model on new data causes catastrophic forgetting.
To address this problem, we formulate fault type classification and fault zone localization as continual learning (CL) problems and design four evaluation scenarios on the IEEE 13-node test feeder, three class-incremental and one domain-incremental. We then propose Prototype-based Dark Experience Replay (ProDER), which extends DER++ with prototype attraction and prototype-level repulsion losses that stabilize the feature space, temperature-scaled logit distillation, and a prototype-aware replay memory that retains both core and boundary samples of each class.
ProDER achieves the highest accuracy among the tested CL methods in all scenarios, with an average accuracy of 58.2\%, 6.6 points above the strongest competing method (DPDMR, 51.6\%) and only 3.2 points below joint training (61.4\%). Per scenario, it improves over the strongest competitor by 4.2 to 7.4 points and closes the gap to joint training to as little as 1.0 point in fault type classification, while matching it in fault zone localization. Moreover, it remains the best method when the replay buffer is substantially reduced.
These results show that prototype-guided replay is an effective, memory-bounded way to keep fault diagnosis models up to date as the grid evolves, while validation on field measurements remains a necessary next step.}
}




\maketitle
\section{Introduction}
\label{intro}

Smart grids integrate real-time monitoring, two-way communication, distributed renewable generation, and intelligent control into the power distribution system~\cite{bayindir2016smart,muqeet2023state,10659889}. As many European grids are being modernized towards this architecture~\cite{RePEc:eee:rensus:v:67:y:2017:i:c:p:776-790}, the amount of measurement data available to operators grows, and so does the variety of operating conditions under which faults occur.

Machine Learning (ML) and Artificial Intelligence (AI) have emerged as powerful tools for enhancing decision-making in smart grid systems. They have been applied to a wide range of tasks, including stability prediction~\cite{9079864,onder2023classification,8454830,EFATINASAB2025101799,ENGLAND2020106189,ALLAL2024108304,EFATINASAB2025101662}, load forecasting~\cite{7926112,https://doi.org/10.1155/2022/4049685,9765492,en15218079,ZHANG2020103852,FONTANACRESPO2024109352}, energy demand management~\cite{9218929,su14052546,10064703,9398860,8299470,CHREIM2023106886} and fault analysis~\cite{10659889,6850055,10081504,su15108348}. Among these, fault diagnosis is one of the most critical, because the faster a fault is identified and located, the faster it can be isolated and service restored~\cite{10659889}.

Throughout this paper, we use the term \textit{fault diagnosis} for the post-fault task of (i) classifying the \textit{fault type} (e.g., line-to-ground, line-to-line, or three-phase faults) and (ii) localizing the \textit{fault zone} from acquired measurements.
Accurate and timely fault diagnosis enables proactive maintenance, minimizes service disruptions, and improves response time, making it indispensable for the next generation of intelligent and self-healing power systems. 

Although many ML and Deep Learning (DL) models have been proposed for fault diagnosis in smart grids, their real-world applicability remains limited. A first obstacle is the scarcity of labeled data: collecting a representative dataset that covers all fault types and all parts of the grid is slow, especially for rare faults, and waiting for such a dataset before deploying a model would leave the grid without automated diagnosis in the meantime. Models should therefore be deployed as soon as a reasonable amount of data is available and then acquire new knowledge over time.

A second obstacle is that smart grids are not static. New fault patterns arise, operating conditions change, and monitoring is extended to additional zones, so models trained on historical data may degrade when faced with novel or shifting conditions~\cite{MAHADEVAN2024111610}.
Retraining from scratch on all old and new data is the usual remedy, but its cost grows with the accumulated data and it requires storing the full history~\cite{verwimp2025accuracytwicefastcontinuous}. Continual Learning (CL) offers an alternative: models are updated incrementally on new data while limiting the forgetting of previously acquired knowledge, with bounded memory and computation.

In this paper, we propose leveraging CL techniques to address the limitations of static models in fault diagnosis for smart grids. To this end, we design a comprehensive evaluation framework that captures the evolving nature of real-world deployments for smart grids through four progressively complex scenarios. 
These are grounded in two core CL paradigms: class-incremental learning, where the model incrementally encounters new fault types or grid zones; and domain-incremental learning, where fault classes remain fixed, but data distributions shift as new fault zones are introduced. 
This setup reflects realistic operational constraints in which models must adapt to new information over time without retraining from scratch. 
{Our evaluation shows that replay-based CL approaches, such as DER, retain considerably more knowledge than regularization-based methods across the considered scenarios.

Building on this, we introduce Prototype-based Dark Experience Replay (ProDER), a replay-based approach that integrates class prototypes into DER. ProDER combines feature-level regularization, through attraction of samples to their class prototype and repulsion between prototypes of different classes, with logit-level distillation and a prototype-aware strategy to populate the replay memory.
Furthermore, we introduce an efficient prototype-level repulsion loss whose cost scales with the number of classes rather than with the batch size (see Sec.~\ref{subsub:loss_computation}).
ProDER achieves the highest accuracy among the tested CL methods in all four scenarios, reducing the gap to joint training to at most 7.8 for fault type classification and matching joint training for fault zone localization.
These findings indicate that CL is a promising approach for keeping fault diagnosis models up to date in evolving grids.

Our contributions are summarized as follows: 

\begin{itemize}
\item To the best of our knowledge, this work provides the first systematic study of fault type classification and fault zone localization in smart grids under a continual learning setting. Unlike conventional static fault diagnosis models, the proposed formulation explicitly considers evolving fault classes and operating conditions.
\item We design and implement four progressively challenging scenarios, each introducing additional complexities to better reflect real-world smart grid fault diagnosis environments.
\item We evaluate the performance of multiple CL strategies across these scenarios, demonstrating their effectiveness under the considered continual learning scenarios for smart grid fault diagnosis.
\item We introduce ProDER, a prototype-guided dark experience replay method for continual fault diagnosis. ProDER jointly addresses representation drift and output drift by combining feature-level prototype regularization, logit-level distillation, and prototype-guided replay memory construction. This design enables more stable adaptation under both class-incremental and domain-incremental changes.
\item The source code used to produce the experiments in this study is publicly available at \url{https://github.com/emadef1/ProDER}.
\end{itemize}

The rest of the paper is organized as follows. Section~\ref{RW} reviews the related work on fault diagnosis in smart grids, continual learning, and prototype-based continual learning. Section~\ref{Method} presents the methodology: the fault type classification and fault zone localization models, the formulation of the continual learning problem, the four evaluation scenarios, and the continual learning methods considered in the comparison. Section~\ref{sec:our_approach} introduces ProDER, describing its distillation loss, the semantic prototype alignment loss, the prototype-aware selection strategy for the replay memory, the motivation behind the design, and the complexity analysis. Section~\ref{sec:ExpSetting} describes the experimental setting, namely the model implementation, the training configuration, the hyperparameters of the compared approaches, the evaluation protocol, and the dataset. Section~\ref{sec:results} reports the results for each scenario, followed by the ablation study, the computational runtime, the robustness to a reduced replay-buffer budget, a discussion of the findings, and the limitations and future directions of this study. Finally, Section~\ref{Con} concludes the paper by summarizing the main findings.

\section{Related Work}
\label{RW}
This section begins with a comprehensive review of existing research on fault diagnosis systems. Subsequently, Section~\ref{CL} will introduce the concept of CL, outlining its core learning paradigms and the principal methodologies developed in this area.

\subsection{Fault Diagnosis in Smart Grids}
According to the categorization presented by Saha et al. \cite{saha2009fault}, fault location strategies in power systems can be broadly divided into three groups: traditional, observant, and intelligent methods. Traditional methods often rely on manual reporting from customers, such as notifying operators about visible damage or unusual smells like burning cables. Observant methods, by contrast, involve automated detection through smart meters or localized sensors that communicate fault events back to the control center. The most sophisticated category, intelligent methods, relies on advanced technologies like smart sensors, artificial intelligence algorithms, and expert systems to detect and classify faults automatically, eliminating the need for human oversight~\cite{en16052280}. This paper centers specifically on this approach, with a particular emphasis on machine learning and AI-driven models for enabling fully automated and adaptive fault diagnosis in smart grid environments.

A wide range of studies have explored various AI techniques for fault detection and classification in smart grids~\cite{en13133460,7398152,en16052280,en13092149,inproceedings,10659889}. For instance, this study~\cite{GHAEMI2022107766} introduces a stacking-based ensemble learning approach for fault type and location identification in smart distribution networks, utilizing multiple base classifiers such as Support Vector Machine (SVM), Random Forest (RF), and K-Nearest Neighbor (KNN). Artificial Neural Networks (ANNs) have been extensively explored in prior research as a means for detecting and predicting faults in power systems~\cite{4441599,doi:10.1049/oap-cired.2017.0007,al2003fault,1413307,aslan2012alternative}, for example, Farias et al .~\cite{FARIAS201820} introduces a cost-effective approach for locating high impedance faults in overhead distribution systems by combining a polynomial voltage-current model with a neural network that is continuously trained after fault occurrence. 
Furthermore, tree-based algorithms such as Random Forests and Gradient Boosting have been widely applied in various studies \cite{s22020458,GHAEMI2022107766,okumus2021random,SAPOUNTZOGLOU2020106254,9817473} for fault detection tasks. 

The use of Recurrent Neural Networks (RNNs) has proven highly effective in many fault detection studies \cite{9072421,SHADI2022107399,8372054} due to their ability to model temporal dependencies and sequential patterns in time-series data. 
Reddy et al.~\cite{6746071} present a smart fault location technique for transmission lines that combines GPS-synchronized current measurements and computational intelligence to accurately identify faults.

While many of the existing works have presented high-performing models, a significant limitation is that they are predominantly trained in a static setting and have not considered the evolving nature of smart grids and real-world implementation problems. 
In particular, they assume that all fault types and all grid zones are represented in the training set from the beginning, and they do not address how the model should be updated when this assumption is violated, which is the problem studied in this paper (see Section~\ref{intro}).

\subsection{Continual Learning}
\label{CL}

Continual Learning aims to incrementally train models on a sequence of tasks.
When visiting each task over time, the model updates itself by adapting to the new task while retaining previously learned knowledge. This process is hindered by catastrophic forgetting, which tends to forget previous knowledge without the specific techniques introduced by CL.

Most of the CL literature considers two CL scenarios: Class-Incremental Learning (CIL) and Domain-Incremental Learning (DIL).
In the CIL scenario, it is expected that each new task introduces new data associated with new classes never been seen previously.
During inference, given a sample, the model needs to predict the correct classes among all the seen ones.
In the DIL scenario, each new task introduces new domains while keeping the output size fixed (same class set) among tasks.

In the CL literature, the methods can be grouped into three big families of approaches known as rehearsal-based, regularization-based, and architecture-based.
\textbf{Rehearsal-based techniques} assume storing and reusing past data samples during training. 
The most well-known method is Experience Replay (ER) \cite{rolnick2019experience}, also known as Replay. 
Replay stores a small portion of old samples in a buffer (or replay memory).
These samples are revisited together with the new samples when visiting a new task.
In this way, the model can retain the accumulated knowledge by periodically revisiting previous data while learning new information. 
\\
\textbf{Regularization-based approaches} consider additional constraints or penalties during training to maintain the memory of old tasks. 
Two representative methods of this category are Elastic Weight Consolidation (EWC) \cite{kirkpatrick2017overcoming} and Learning without Forgetting (LwF) \cite{li2017learning}.
EWC associates with each model weight a value that indicates how important that weight is for previously seen tasks, and during training a regularization term makes it more difficult to change the important weights.
\\
Instead, LwF is based on the concept of \textit{Knowledge Distillation}, introduced by Hinton et al.~\cite{hinton2015distilling}.
Basically, the current model tries to produce the same output (logits) that would be produced by the model from the previous task.
This approach encourages the model to produce consistent outputs across both old and new tasks, helping it retain prior knowledge while learning new information. 
\\
\textbf{Architecture-based approaches} modify the original model's architecture to preserve existing knowledge. The specific methods used in these approaches vary widely \cite{rusu2016progressive, fernando2017pathnet, mallya2018packnet}. However, these methods generally have two main limitations: they require memory that increases with the number of tasks, and they often necessitate the task ID during inference. These factors render them unsuitable for our objectives.

\subsection{Prototype-based Continual Learning}
\label{sec:proto_rw}
Some works have explored the use of prototypes in the continual learning context.
A prototype refers to a representative vector, typically the mean embedding of samples belonging to a given class, that summarizes the class distribution in the learned feature space and serves as a compact abstraction for learning and memory.
Existing methods differ mainly in \textit{where} the prototype enters the learning pipeline: as the classifier, as an anchor in the loss, or as a criterion to select the replay memory. 

\textbf{Prototypes as classifier.} iCaRL~\cite{rebuffi2017icarl} combines a replay memory with a nearest-mean-of-exemplars classifier and selects exemplars by herding, i.e., it keeps the samples that best approximate the class mean. Its memory therefore concentrates on the class core and discards boundary samples, which are the ones that carry information about class margins.
Moreover, its distillation acts on the outputs only and does not constrain the geometry of the feature space.
CoPE~\cite{De_Lange_2021_ICCV} also classifies by nearest prototype, but it is designed for \textit{online} data-incremental learning, where each sample is seen once and prototypes are updated with momentum. Aghasanli et al.~\cite{Aghasanli_2025_CVPR} use a nearest-prototype classifier on top of a frozen, large pre-trained vision backbone and store a label-free buffer of latent representations. In all three cases, the linear output layer is replaced by a distance-based classifier, which removes the logits on which DER-style distillation relies.

\textbf{Prototypes in the loss.} PRD~\cite{pmlr-v202-asadi23a} learns prototypes in a supervised-contrastive latent space and distills prototype--sample relations to avoid storing past data. It targets the replay-free regime and relies on contrastive representation learning with image augmentations, for which no established counterpart exists for the multivariate electrical measurements considered here. DPDMR~\cite{10719982} combines a prototype compactness loss, a triplet loss, and a K-Means-based prototype-guided memory, and was developed for acoustic (one-dimensional) signal classification and align closely with our setting in terms of data modality and training protocol.

\textbf{Prototypes for memory selection.} PMR~\cite{10058177} uses prototypes to select the samples stored in memory within an online meta-learning framework for text classification. Like iCaRL and DPDMR, it uses prototypes to decide \textit{which} samples to store, but not to shape the information stored with them.

Compared to these works, our approach, ProDER differs in three respects. First, it keeps the standard linear classifier and uses prototypes only as anchors in the loss, so that prototype regularization and logit distillation act on the same network: the attraction loss shapes the features, while distillation preserves previously learned output relationships that, under the approximate conditions discussed in Proposition~\ref{prop:geometry}, can reflect prototype-distance structure. No existing method couples prototype regularization with Dark Experience Replay in this way. Second, ProDER stores both core and boundary exemplars, rather than only the samples closest to the prototype (iCaRL, PMR) or to K-Means centers (DPDMR). Third, it enforces inter-class separation at the prototype level, with a pairwise cost that scales with the number of classes instead of the number of samples, as in triplet or contrastive losses.

\section{Methodology}
\label{Method}

We begin by describing the fault type classification and fault zone localization models and the dataset used in this work. We then formalize the CL problem and introduce the four scenarios used to evaluate adaptation under evolving fault classes and grid zones. Finally, we describe the CL approaches considered in the evaluation.

\subsection{Fault Type Classification and Fault Zone Localization Models}
\label{SM}
In this section, we discuss the core functionality of fault type and fault zone classification models within a smart grid environment. These models operate as two distinct yet interrelated components, both leveraging the same input data sourced from various devices deployed across the grid, such as sensors, smart meters, and other monitoring equipment.

We use the IEEE13-AdvAttack dataset~\cite{mainpaper}, which, to the best of our knowledge, is one of the few publicly available datasets with labeled fault events for both fault type and fault zone; we are not aware of a public dataset of real field measurements with comparable annotations. The data are simulated on the IEEE 13-node test feeder, a 4.16~kV distribution network whose 13 buses are instrumented to record three-phase voltages and currents. The feeder is partitioned into four zones, around buses 671, 633, 675, and 680, and in each zone 11 fault types are injected. Each observation is described by 51 features, which include measurements from both conventional grid components and renewable sources, under normal and faulty operating conditions.

The input data consists of both time-domain and frequency-domain features. Time-domain features include statistical descriptors such as mean, standard deviation, skewness, kurtosis, signal energy, and peak amplitude, capturing the temporal dynamics of voltage and current signals. In parallel, frequency-domain features are extracted using signal processing techniques like the Discrete Fourier Transform (DFT) and the Discrete Wavelet Transform (DWT), which enable multi-resolution analysis of transient and oscillatory components in the signals. 

The \textbf{fault type classification model} is responsible for classifying the nature of the fault (phase to ground, phase to phase, phase to phase to ground, three phase, three phase to ground, for a total of 11 faults) based on these extracted features. 
Simultaneously, the \textbf{fault zone localization model} aims to localize the fault by identifying the specific region (4 zones) or segment within the power distribution network where the fault has occurred. Accurate zone localization is critical for enabling fast fault isolation, efficient dispatch of maintenance teams, and minimizing service disruption.

Both models are trained with CL techniques using historical and simulated fault data from the grid. Once deployed, they operate in real-time, continuously processing incoming data and providing predictive insights with high accuracy, thereby enhancing the resilience and reliability of smart grid operations.

\subsection{Problem formulation}
Fault diagnosis in sensor-based systems is essential for ensuring operational reliability and minimizing downtime in real-world industrial applications. These systems often encounter new types of faults or evolving fault characteristics over time, making static models ineffective. CL provides a scalable solution by enabling models to incrementally learn from a stream of tasks while retaining previously acquired knowledge.

In our setting, the model is trained over a sequence of $T$ tasks. Each task $t$ corresponds to either a \textit{fault type classification} or a \textit{fault zone localization} problem and is associated with a dataset:
\begin{equation}
\mathcal{D}_t = \{ \mathbf{X}_t, \mathbf{Y}_t \}
\end{equation}
where:
\begin{itemize}
    \item $\mathbf{X}_t = [\mathbf{x}_{1t}, \dots, \mathbf{x}_{n_t t}]$, with $\mathbf{x}_{it} \in \mathbb{R}^{W \times F}$ representing a multivariate time-series segment (i.e., sensor values over $W$ time steps and $F$ features),
    \item $\mathbf{Y}_t = [y_{1t}, \dots, y_{n_t t}]$, where each label $y_{it} \in \mathcal{C}_t$ belongs to the task-specific class set.
\end{itemize}

We consider two separate CL Problems:
\begin{itemize}
    \item \textbf{Fault Type Classification:} The label space is $\mathcal{C}^{(\text{type})} = \{c_1^{(\text{type})}, \dots, c_{11}^{(\text{type})}\}$, and tasks are defined over disjoint subsets $\mathcal{C}_t^{(\text{type})} \subset \mathcal{C}^{(\text{type})}$.
    
    \item \textbf{Fault Zone Localization:} The label space is $\mathcal{C}^{(\text{zone})} = \{c_1^{(\text{zone})}, \dots, c_4^{(\text{zone})}\}$, with tasks defined over subsets $\mathcal{C}_t^{(\text{zone})} \subset \mathcal{C}^{(\text{zone})}$.
\end{itemize}

For each task $t$, the model $f_{\theta_t}$ is trained to learn the mapping:
\begin{equation}
f_{\theta_t}: \mathbf{x} \rightarrow y 
\quad \text{where} \quad 
y \in \bigcup_{i=1}^{t} \mathcal{C}_i
\label{eq:model-mapping}
\end{equation}
with the objective of maintaining performance across all previously seen classes (i.e., minimizing forgetting) while learning new ones.

For the domain-incremental Scenario~3 (Section~\ref{sc}), the label set is the same for every task, $\mathcal{C}_t=\mathcal{C}^{(\text{type})}$, and tasks differ only in the input distribution $P_t(\mathbf{x})$, since each task contains data from a different grid zone. In all scenarios, the task identity is not available at inference time. The model is updated task by task, with multiple training epochs on each task; online learning from a single pass over a data stream is not considered in this work.

The goal is to develop and evaluate CL strategies that achieve robust performance on both fault-type and fault-zone classification \textit{without access to the full training data of prior tasks}: when training on task $t$, the learner can only access $\mathcal{D}_t$ and, for replay-based methods, a memory buffer $\mathcal{M}$ of fixed capacity $M$ that stores a subset of previously seen samples and does not grow with the number of tasks. This bounded-memory assumption is what distinguishes CL from retraining on the accumulated data.

\subsection{Scenarios}
\label{sc}
In this section, we propose four realistic CL scenarios tailored to the dual task of fault type and fault zone localization within smart grids. These scenarios are designed to reflect the dynamic nature of real-world grid environments, where new types of faults may emerge, and fault occurrences may shift to previously unseen zones of the network. 

The objective across all scenarios is to train models that can progressively adapt to new fault patterns or new grid zones while preserving their ability to accurately identify previously learned fault types and localizations. This ensures that the model maintains comprehensive and up-to-date fault coverage without suffering from catastrophic forgetting, making it suitable for long-term deployment in evolving smart grid infrastructures.

\subsection{Scenario 1 – Fault Type Classification with Two New Fault Classes (Class-Incremental)}
In this scenario, we simulate a realistic deployment cycle for a fault type classification system in a smart grid environment. Initially, the system is trained using a labeled dataset containing three fault classes (classes 0–2). The decision to begin with three classes is based on a practical assumption as identifying more than two distinct fault types is sufficient to ensure the basic operational effectiveness of the system in its early stages. This allows the system to be deployed promptly while covering a meaningful range of fault conditions. 

Once deployed, the system continues to collect operational data. Over time, sufficient labeled samples become available for two previously unseen fault classes. To maintain predictive performance and accommodate these new classes, the model must be updated. However, retraining from scratch is inefficient and risks forgetting previously learned knowledge, making this a natural fit for a CL approach. 

From a CL standpoint, we structure the dataset as five sequential tasks: the first task includes samples from the initial three fault classes, while each of the remaining tasks introduces data from two new classes (see Figure~\ref{fig:sample1}). This setup aligns with the class-incremental learning paradigm discussed in Section~\ref{CL}, where the model must learn new classes over time while retaining its ability to classify previously seen ones. This setup mirrors the conventional class-incremental learning scenario and provides a controlled yet realistic benchmark for evaluating CL capabilities in fault classification systems.

\begin{figure}[!htbp]
  \centering
  \includegraphics[width=\textwidth]{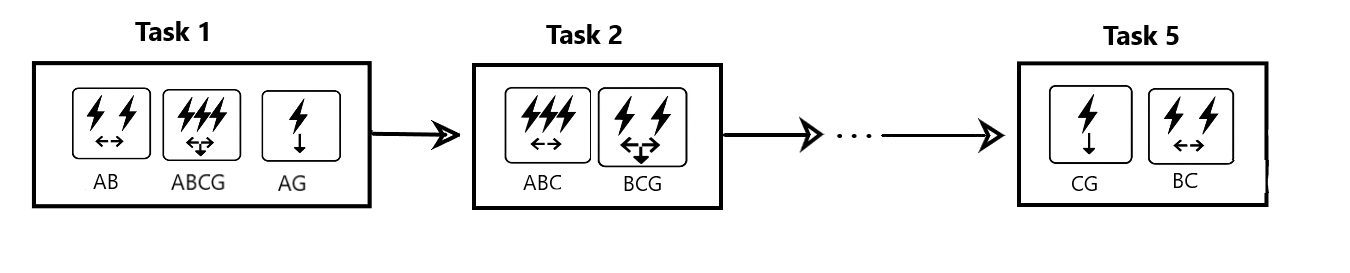}
  \caption{\textbf{Task sequence for Scenario 1: The first task includes 3 fault types, and each of the remaining 4 tasks introduces 2 new fault types, covering all 11 faults.}}
  \label{fig:sample1}
\end{figure}

\subsection{Scenario 2 – Fault Type Classification with One New Fault Class (Class-Incremental)}
This scenario represents a more granular and challenging CL setting for fault type classification. As in Scenario 1, we begin by training a model using labeled data for an initial set of fault types (0-2). However, unlike the previous case, where two classes were introduced after the first task, here we take a more incremental approach by presenting only one fault class per task. Therefore, we divide the dataset into nine sequential tasks (see Figure~\ref{fig:sample2}).

The motivation behind this design is to enable quicker integration of new data into the operational system. Rather than waiting to accumulate data for multiple fault types, the system is updated as soon as data for a single new class becomes available. This allows for faster adaptation in real-world deployments where fault data may arrive asynchronously or sporadically. 

From a CL perspective, this setup is significantly more challenging. Although the total number of fault classes remains the same as in Scenario 1, the dataset is now divided into nine sequential tasks instead of five. Each task introduces a single new class, requiring more frequent updates to the model. This increases the risk of catastrophic forgetting, as the model is exposed to fewer examples per task and must revise its internal representations more often.

Moreover, learning classes individually, rather than in batches, can hinder the model’s ability to form distinct and well-separated representations of each fault type. With limited inter-class context during each training phase, the model may struggle to generalize across fault boundaries, making the learning process more susceptible to interference.
This scenario closely aligns with the class-incremental learning paradigm discussed in Section~\ref{CL} and serves as a rigorous benchmark to assess the robustness and stability of CL strategies under more realistic and fragmented data acquisition conditions.

\begin{figure}[!htbp]
  \centering
  \includegraphics[width=\textwidth]{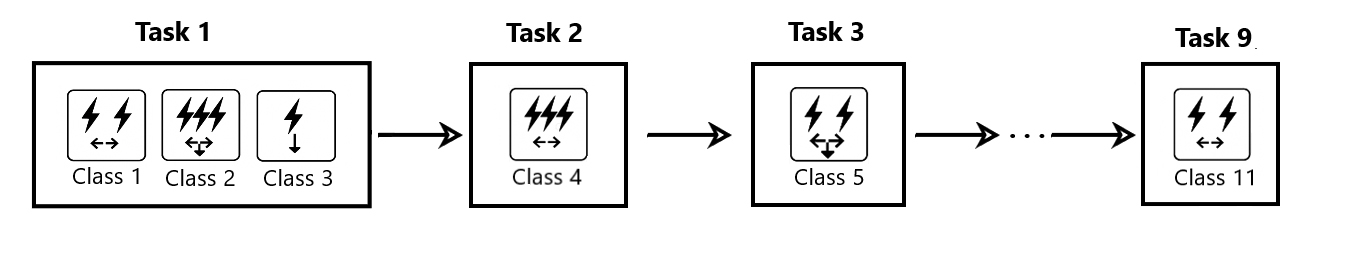}
  \caption{\textbf{Task sequence for Scenario 2: The first task includes 3 fault types, and each of the remaining 8 tasks introduces 1 new fault type, covering all 11 faults.}}
  \label{fig:sample2}
\end{figure}

\subsection{Scenario 3 – Fault Type Classification with Known Faults in a New Grid Zone (Domain-Incremental)}
In this scenario, we focus on the generalization capabilities of a fault type classification model when deployed in geographically or structurally distinct parts of the smart grid. Rather than introducing new fault types, we retain the same set of fault classes across all tasks. The variation instead comes from the input data, which is drawn from different grid zones over time.

The scenario begins with the model trained on fault data from the first initial zone. This data includes labeled examples of known fault types, allowing the model to learn patterns and characteristics associated with those faults within a specific operational context. After deployment, as the system expands or is rolled out in new areas of the grid, labeled data from additional zones becomes available. Each new task introduces data from a new grid zone while maintaining the same set of fault type labels (see Figure~\ref{fig:sample3}).

From a CL standpoint, this setup constitutes a domain-incremental scenario. The classification task remains unchanged (i.e., predicting the same fault types), but the input distribution shifts due to differences in grid conditions and sensor behaviors across zones. The challenge for the model is to adapt to the new domains without degrading performance on previously seen ones. We divide the dataset into four sequential tasks, with each task including data from an unseen zone. Unlike the previous scenarios, which fall under the CIL setting, this scenario follows the domain-incremental (DIL) setting described in Section~\ref{CL}. 

This scenario is particularly relevant in real-world smart grid deployments, where infrastructure heterogeneity is common. A model trained in one zone may not generalize well to another without domain adaptation. By evaluating CL methods in this setting, we can assess their ability to retain accuracy across shifting domains, making this scenario a crucial benchmark for building robust and transferable fault diagnosis systems.  

\begin{figure}[!htbp]
  \centering
  \includegraphics[width=\textwidth]{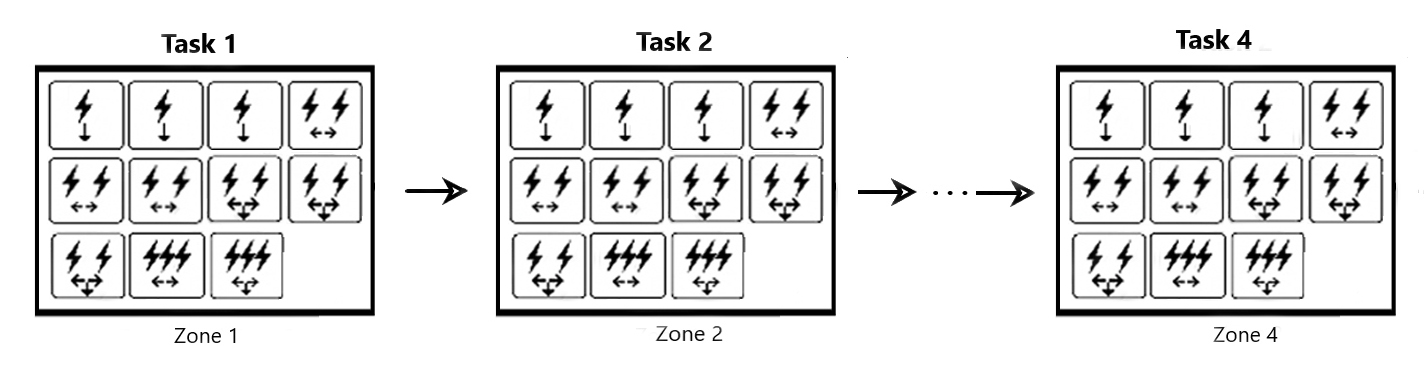}
  \caption{\textbf{Task sequence for Scenario 3: All tasks contain the same set of 11 fault types, but each task presents data from a different zone or domain. As a result, the next task expand the knowledge of previously encountered classes.}}
  \label{fig:sample3}
\end{figure}

\subsection{Scenario 4 – Fault Zone Localization with New Zone Labels (Class-Incremental)}

In this scenario, we shift the focus from predicting fault types to localizing the fault’s origin within the smart grid, i.e., fault zone localization. The task involves identifying the specific region or segment of the grid where a fault has occurred, which is critical for timely diagnostics, response coordination, and minimizing service disruptions.

Initially, the model is trained using labeled data corresponding to a subset of known fault zones within the grid. These zones are treated as distinct output classes. Once the system is deployed, it operates in real-time to predict the zone associated with each detected fault. Over time, as the grid infrastructure evolves or monitoring coverage expands, data from additional zones becomes available. These new zones correspond to previously unseen class labels, requiring the model to update its knowledge and expand its classification space.

From a CL perspective, this scenario represents a class-incremental setting. Each new task introduces a new zone label, challenging the model to integrate new classes while maintaining its ability to correctly identify zones it has already learned. To simulate this process, we partition the dataset into three sequential tasks. The first task includes labeled data from the first two grid zones, task 2 introduces data from a third zone, and task 3 adds data from a fourth, unseen zone(see Figure~\ref{fig:sample4}).  As in other class-incremental setups, the model must avoid catastrophic forgetting while preserving and extending its classification capabilities. 

\begin{figure}[!htbp]
  \centering
  \includegraphics[width=\textwidth]{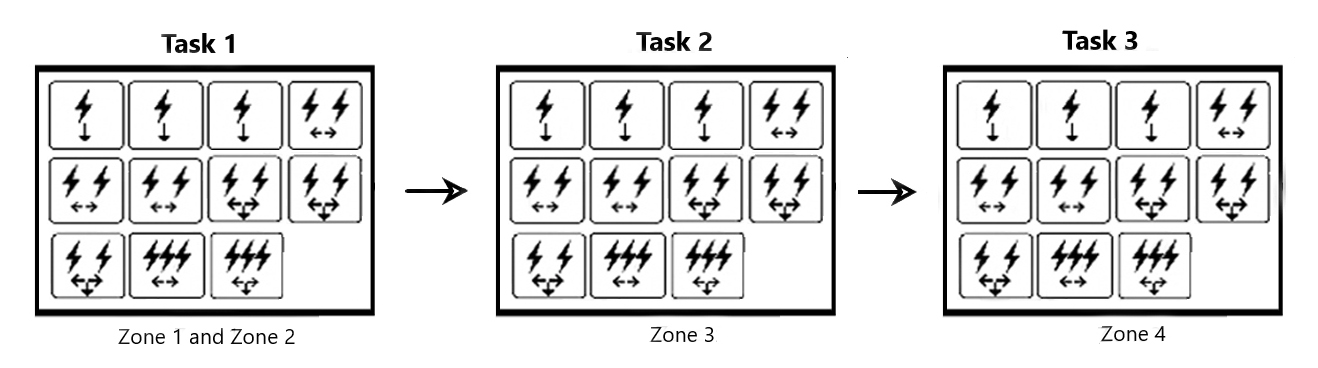}
  \caption{\textbf{Task sequence for Scenario 4: The first task includes 2 fault zone classes, and each of the following 2 tasks introduces 1 new fault zone class, resulting in 3 tasks covering all fault zones. As a result, each new task expands the knowledge of previously encountered classes. }}
  \label{fig:sample4}
\end{figure}

\subsection{Continual Learning Methods}
\label{CL_method}
In this section, we present the CL methods evaluated in this study. We include two baselines to establish performance bounds: Fine-Tuning as a lower bound, and Joint Training as upper bound. In addition to these, we assess several widely adopted CL approaches, including Experience Replay (ER), Elastic Weight Consolidation (EWC), Learning without Forgetting (LwF), Disentangled Prototype-guided Dynamic Memory Replay (DPDMR) and Dark Experience Replay (DER). Finally, we introduce our proposed CL technique, referred to as ProDER, which is specifically designed for the fault diagnosis setting explored in this work.
For a fair comparison among replay-based methods, we use the same replay mechanism and memory budget (23.7\% of training data) whenever applicable. The construction of the replay memory, however, follows the strategy defined by each method. ER uses its standard random memory update, DER++ follows its replay-buffer formulation, DPDMR applies its prototype-guided memory selection, and ProDER uses the proposed prototype-aware exemplar selection. This keeps the replay conditions comparable while preserving the method-specific memory management strategy of each approach. The domain-incremental scenario undergoes slightly different preprocessing, resulting in a marginally different number of training samples and, consequently, a slightly different memory size; nevertheless, the replay memory in all scenarios is limited to no more than 23.7\% of the corresponding training data.
\begin{itemize}
    \item Joint Training (Offline Reference): This method assumes access to the entire dataset across all tasks from the beginning. The model is trained once on the full dataset. Although not a realistic scenario, it assumes the best conditions, where all data containing all conditions can be learned simultaneously.

    \item Fine-Tuning: In Fine-Tuning, the model learns each task sequentially without any mechanism to retain knowledge from previous tasks. This naive approach leads to severe catastrophic forgetting and therefore serves as a practical lower bound for CL methods.
    \item EWC~\cite{kirkpatrick2017overcoming}: EWC mitigates forgetting by identifying important model parameters for previous tasks using the Fisher Information Matrix, and penalizing their updates during new task learning. A regularization term weighted by a hyperparameter $\lambda$ (tuned to 10 in our setup) constrains critical weights, preserving prior task knowledge while adapting to new data.

    \item LwF~\cite{li2017learning}: LwF is a distillation-based method that preserves knowledge from previous tasks by encouraging the current model to remain consistent with a frozen copy of its previous state. In our implementation, the distillation term is computed using temperature-scaled Kullback--Leibler (KL) divergence between the output distributions of the current and previous models over the previously learned classes. This term is combined with the classification loss of the current task. The distillation strength is controlled by $\lambda=1$.  The temperature is set to $T=2$ in Scenarios 1 and 3 and $T=1$ in Scenarios 2 and 4.

    \item ER~\cite{rolnick2019experience}: ER combats forgetting by maintaining a fixed-size memory buffer of samples from previous tasks. At each training step, the model is updated using a mixture of new task data and randomly selected replay samples. To balance memory usage across tasks and classes, the buffer is updated by removing samples uniformly. We set the memory size to 363 for scenario 1, 2 and 4 and 353 for scenario 3 (23.7\% of training data in all scenarios).

    \item DER~\cite{buzzega2020darkexperiencegeneralcontinual}: DER extends ER by storing not just the raw input and labels, but also the model’s past output logits (“dark knowledge”). This enables a softer form of memory rehearsal using output-level supervision. DER requires the same memory size and replay ratio as ER but introduces logits-based alignment to enhance retention. An advanced variant of DER, DER++, stores both ground truth labels and the output logits for each sample in the memory. This dual supervision enables more effective retention of prior knowledge. In our experiments, we maintained a memory size of 363 and 353 like the ER case. The balance between the logit-matching and replay classification losses is controlled by the hyperparameters $\alpha$ and $\beta$, respectively. Specifically, we use $(\alpha,\beta)=(4,1)$ for Scenario 1 , $(\alpha,\beta)=(6,1)$ for Scenario 2, $(1,0.5)$ for Scenario 3, and $(2,1)$ for Scenario 4.

    \item DPDMR~\cite{10719982}: DPDMR uses prototype-guided replay to preserve representative samples from previously learned classes. In addition to the classification loss, it employs a prototype compactness loss to keep samples close to their corresponding prototypes and a triplet loss to improve the separation between classes. Multiple prototypes are obtained through K-Means clustering, and samples located close to the resulting cluster centers are selected for replay. The memory is dynamically updated as new tasks arrive while respecting a fixed overall memory budget. For all four scenarios, we used a memory size of 363 for scenario 1, 2 and 4 and 353 for scenario 3. The triplet margin was set to 1.0, while the prototype and triplet losses were both assigned a weight of 1.0. The memory-selection parameters were adjusted according to the task structure of each scenario. Specifically, we used $n_k=15$ and 5 K-Means clusters for Scenario~1, $n_k=10$ and 4 clusters for Scenario~2, $n_k=15$ and 6 clusters for Scenario~3, and $n_k=15$ and 8 clusters for Scenario~4.

\end{itemize}

Choice of baselines. Among the prototype-based methods reviewed in Section 2.3, only DPDMR is included in the comparison. We selected baselines that can be applied to our benchmark without altering the conditions of Section 3.2, namely: (i) offline, task-based training with multiple epochs per task; (ii) no task identity at inference time; (iii) a recurrent backbone trained from scratch on multivariate electrical measurements, without pre-trained models or domain-specific data augmentation; and (iv) a replay memory with a fixed budget. DPDMR satisfies all four conditions and is run with the same backbone, memory budget, replay ratio, and tuning budget as the other replay methods.

The other prototype-based methods violate at least one condition. iCaRL~\cite{rebuffi2017icarl} replaces the linear classifier with a nearest-mean-of-exemplars classifier, removing the logits used by DER-style distillation; CoPE~\cite{De_Lange_2021_ICCV} targets online, single-pass learning; PMR~\cite{10058177} is an online meta-learning method for text; PRD~\cite{pmlr-v202-asadi23a} is replay-free and relies on contrastive learning with image augmentations; and Aghasanli et al.~\cite{Aghasanli_2025_CVPR} build on a frozen, large pre-trained vision backbone. Adapting them would require replacing their classifier, training protocol, backbone, or augmentation pipeline, and the resulting comparison would evaluate our adaptation rather than the original methods. A dedicated adaptation of these approaches to sensor-based time series is left for future work.

\begin{figure}[thbp]
  \centering
  \includegraphics[width=0.9\linewidth]{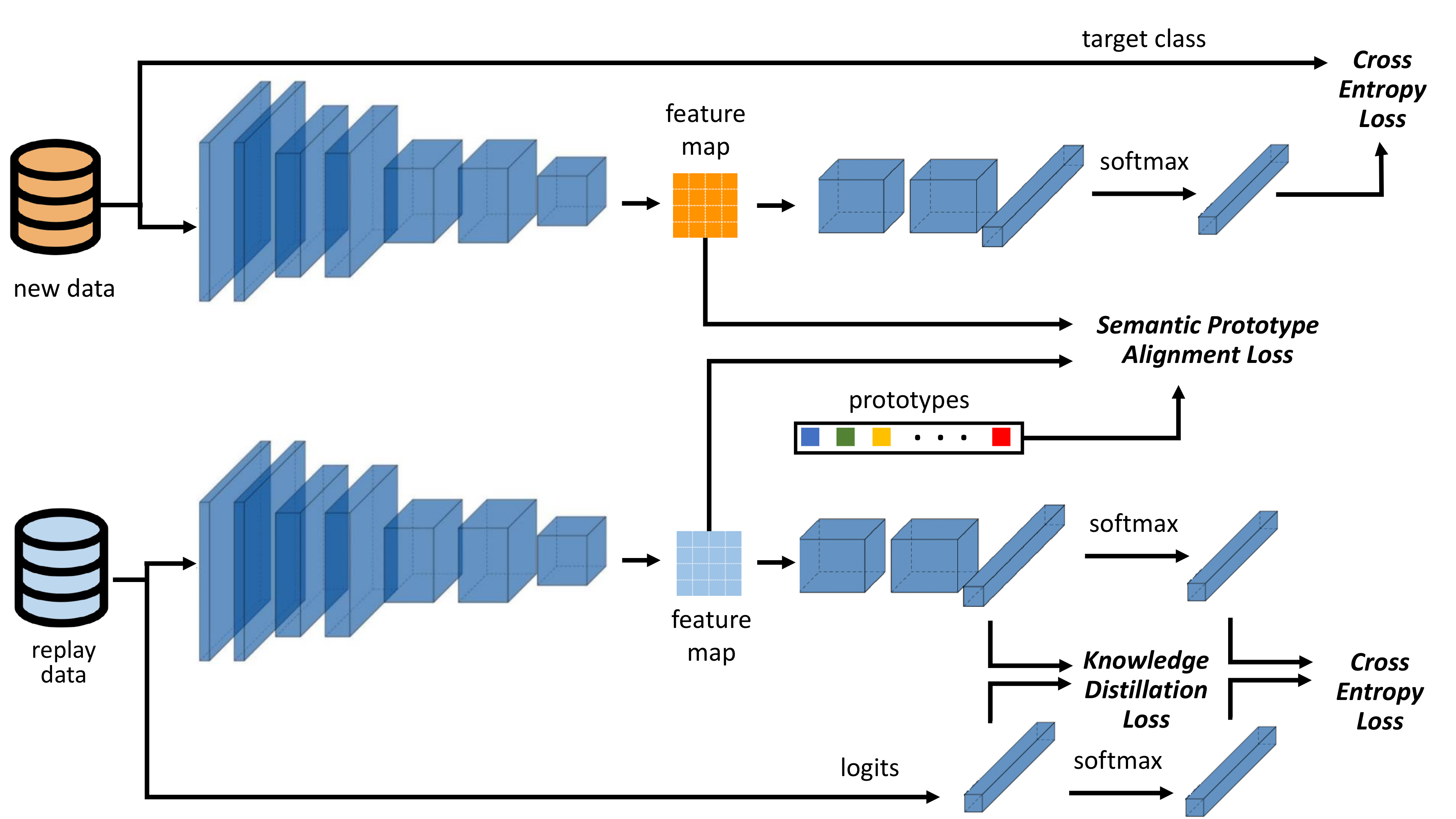}
  \caption{\textbf{ProDER's scheme}. ProDER builds on DER with a multi-objective optimization to reinforce feature stability and class separability across tasks through feature-level regularization, while also employing logit-level distillation and a prototype-aware sample strategy to populate the replay memory.
  }
  \label{fig:ProDER_scheme}
\end{figure}

\section{ProDER: Our Proposed Approach}
\label{sec:our_approach}

We introduce \textbf{ProDER}, a prototype-driven experience replay framework for class-incremental and domain-incremental continual learning, in which the task identity is not available at inference time (the generic scheme of ProDER is illustrated in Figure \ref{fig:ProDER_scheme}). Built upon the foundation of DER++, ProDER integrates prototype-based supervision and a hybrid memory pruning strategy to improve both representation stability and generalization across tasks.

\subsection{Knowledge Distillation Loss}
{DER++ forms the replay-based backbone of our approach by preserving the responses of previously observed samples. In its original formulation, DER++ regularizes the model by matching the stored pre-softmax logits using a squared-error loss, together with experience replay. In ProDER, we modify this response-preservation mechanism by replacing the raw-logit squared-error objective with temperature-scaled knowledge distillation, which provided better knowledge retention in our setting. In addition, rather than employing separate cross-entropy terms for current and replay samples, we compute a single classification loss over the combined current-replay batch.}
Given a model's output $\mathbf{z}$ and the stored logits $\mathbf{z}^{old}$, the distillation loss is computed as:

\begin{equation}
\mathcal{L}_{\text{distill}} =
\mathrm{KL}\left(
\mathrm{softmax}\left(\frac{\mathbf{z}^{old}}{T}\right)
,\middle|,
\mathrm{softmax}\left(\frac{\mathbf{z}}{T}\right)
\right),
\end{equation}

where $T$ is a temperature hyperparameter. This loss encourages the model to maintain similar predictions for past samples, helping mitigate catastrophic forgetting.

\begin{figure}[thbp]
  \centering
  \includegraphics[width=0.9\linewidth]{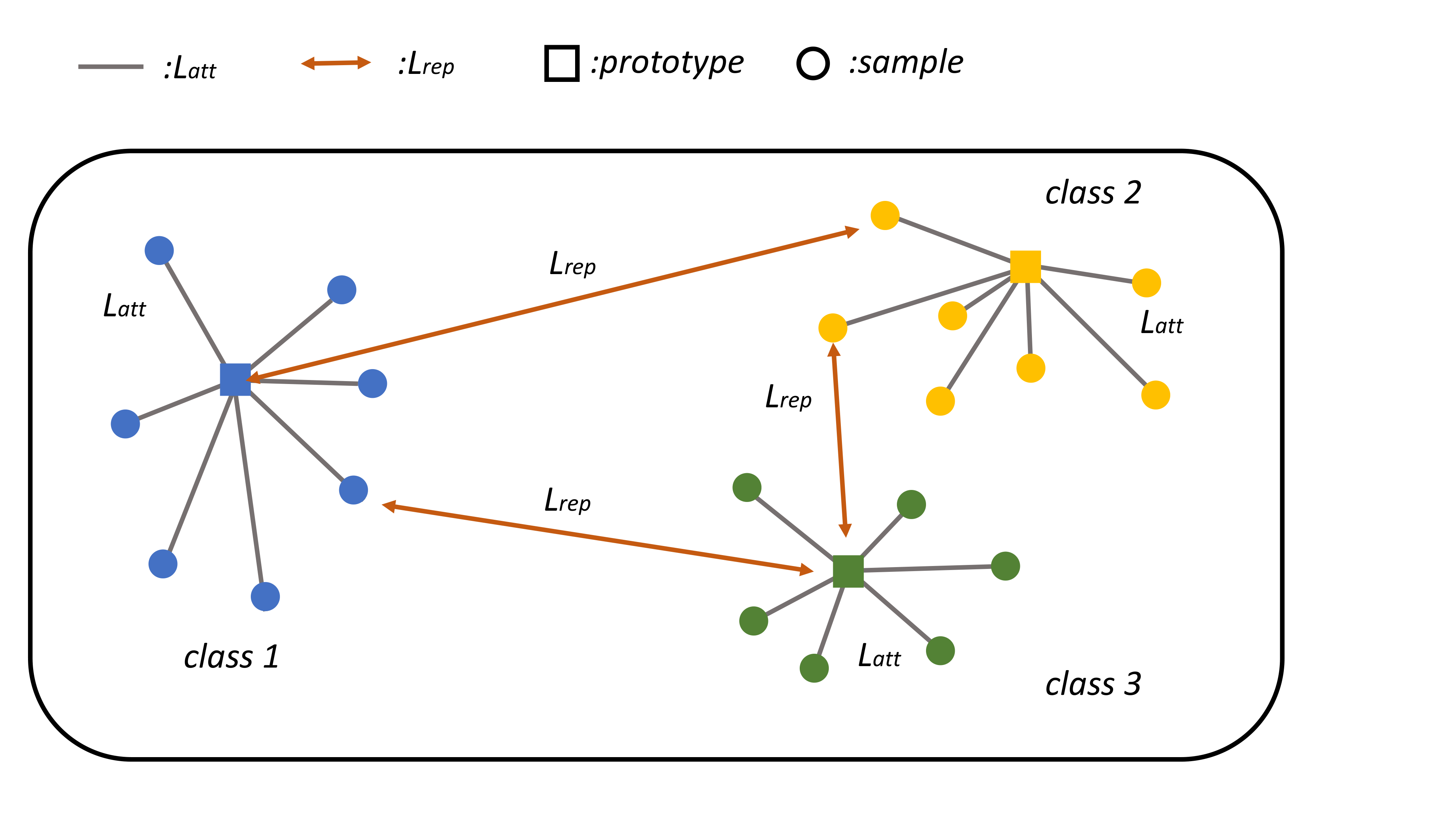}
  \caption{\textbf{Semantic Prototype Alignment (SPA Loss)}. Illustration of the proposed scheme designed to improve the stability of the DER approach by incorporating prototypes. It applies a feature-level regularization via attraction of the samples to class prototypes and repulsion from others.
  This enhances representation stability and class separability across tasks. $L_{att}$ is the attraction loss and $L_{rep}$ represents the repulsion loss.  
}
  \label{fig:Semantic_Prototype_Alignment}
\end{figure}

\subsection{Semantic Prototype Alignment (SPA) Loss}
While methods such as DER and DER++ alleviate forgetting via experience replay and logit distillation, they lack mechanisms to explicitly structure the feature space, making them vulnerable to representation drift, the progressive shift of feature embeddings for old classes during new task training.

ProDER addresses this limitation through a prototype-guided regularization strategy. While the concept of prototypes is not new in CL~\cite{pmlr-v202-asadi23a,Aghasanli_2025_CVPR,De_Lange_2021_ICCV,10058177}, ProDER offers a cohesive and finely tuned framework that embeds prototype guidance into DER++.  ProDER fuses prototype-level regularization (pulling examples toward their class centroids and pushing different centroids apart) with DER++’s soft-target logit distillation, a prototype-aware sampling strategy, and a multi-objective optimization schedule. The result is a coordinated mechanism where (i) the semantic losses produce compact, well-separated class embeddings, (ii) exemplars are selected with respect to those embeddings to maximize replay utility, and (iii) distillation preserves prior decision boundaries. 

At the core of ProDER is the use of class prototypes, empirical mean embeddings of features belonging to each class, as stable anchors in the representation space. 
Specifically, ProDER introduces the Semantic Prototype Alignment (SPA) Loss, which is composed of (i) an attraction loss and (ii) a repulsion loss (see Figure \ref{fig:Semantic_Prototype_Alignment}).

For each class $c$ represented in the current mixed mini-batch, a prototype vector $\mathbf{p}_c$ is computed as the mean feature embedding of the samples belonging to that class:

\begin{equation}
\mathbf{p}_c =
\frac{1}{|\mathcal{B}_c|}
\sum_{i \in \mathcal{B}_c}
\mathbf{f}_i,
\end{equation}

where $\mathbf{f}_i$ is the representation of sample $i$, obtained from an intermediate layer of the network, and $\mathcal{B}_c$ denotes the subset of samples in the current mixed mini-batch that belong to class $c$. Since the mini-batch contains both current-task and replay samples, the resulting prototypes reflect the class representations available during each training step. These prototypes are then used as semantic reference points to encourage compact class representations while the model learns new tasks.

To enforce intra-class compactness, ProDER minimizes the squared distance between each feature vector and its corresponding prototype. We define the \textbf{prototype attraction loss}:

\begin{equation}
\mathcal{L}_{\text{att}} = \frac{1}{N} \sum_{i=1}^{N} \left\| \mathbf{f}_i - \mathbf{p}_{y_i} \right\|^2.
\end{equation}

This loss promotes semantic consistency within each class, improving feature stability across tasks. This loss minimizes the conditional variance $\text{Var}[\mathbf{f} \mid y = c]$ across tasks. A smaller intra-class variance implies more robust, class-specific clustering in the feature space. 
To complement the attraction term, we use a \textbf{repulsion loss} to keep the prototypes of different classes sufficiently separated. Because prototype magnitudes can change during training, each prototype is first normalized as

\begin{equation}
\bar{\mathbf{p}}_c =
\frac{\mathbf{p}_c}
{\|\mathbf{p}_c\|_2 + \epsilon},
\end{equation}

where $\epsilon$ is a small constant added for numerical stability. The repulsion loss is then computed over the unique pairs of class prototypes present in the current mixed mini-batch:

\begin{equation}
\mathcal{L}_{\text{rep}}
=
\frac{1}{|\mathcal{P}|}
\sum_{(i,j)\in\mathcal{P}}
\left[
\max
\left(
0,\,
m -
\left\|
\bar{\mathbf{p}}_i -
\bar{\mathbf{p}}_j
\right\|_2
\right)
\right]^2,
\end{equation}

where $\mathcal{P}$ is the set of unique prototype pairs and $m$ defines the desired minimum separation between them. Since the prototypes are $\ell_2$-normalized, their Euclidean distance is bounded between 0 and 2, which makes the margin independent of the original feature scale. The loss is active only when two prototypes are closer than $m$; once the required separation is reached, no additional repulsion is applied. The margin $m$ is selected per scenario as described in Section~\ref{sec:ExpSetting}.

Together, these two objectives impose a structured regularization on the learned feature space. The attraction loss acts as a pulling force, encouraging individual features to align tightly with their class center, thus enhancing intra-class compactness. Meanwhile, the repulsion loss exerts a pushing force between class prototypes, ensuring that different classes remain well-separated. This dual mechanism leads to clearer cluster boundaries and a more stable representation, which is especially important in CL scenarios where feature drift and class confusion are common. As a result, ProDER achieves a balance between class cohesion and discriminability, facilitating better retention of prior knowledge and smoother integration of new tasks.

\subsection{Prototype-aware Selection Strategy}
While prototype-based regularization governs training dynamics, sample efficiency in replay memory remains a key bottleneck. Existing methods, such as DER++, store logits and randomly sample exemplars, which often leads to semantically redundant or unrepresentative buffers. This randomness becomes particularly detrimental when replay memory is small, like in our fault diagnosis case, where each sample's quality plays a crucial role in mitigating forgetting.
ProDER addresses memory management by proposing a \textbf{prototype-aware selection strategy} (see Figure \ref{fig:prototype_aware_selection_strategy}). The goal is to retain samples that are both \textit{representative}, close to the prototype and central to class semantics, and \textit{diverse}, far from the prototype, capturing the variation within the class. Let the memory buffer for class $c$ be denoted as
\begin{equation}
\mathcal{M}_c = \{ (x_i, y_i, z_i,f_i) \}
\label{eq:class-memory}
\end{equation}
where $x_i$ is the input sample, $y_i$ is the class label, $z_i$ is the model's output logits.
Let $f_i$ be the learned feature representation of the sample (typically extracted from a penultimate layer). For class $c$, the distance to the class prototype is computed for each sample:

\begin{equation}
d_i = \left\|\mathbf{f}_i - \mathbf{p}_c\right\|^2.
\end{equation}

Samples are then ranked by their distance to the class prototype, denoted as \( d_i \).
We define $\rho \in [0, 1]$ as the ratio controlling the trade-off between representative and diverse samples in the hybrid memory selection strategy. 
The selection process retains the \( \lfloor \rho K \rfloor \) closest samples (representative instances) and the remaining \( K - \lfloor \rho K \rfloor \) farthest samples (diverse instances), forming the selected index set:

\begin{equation}
\mathcal{I}_c =
\text{Bottom}_{\lfloor \rho K \rfloor}(d)
\cup
\text{Top}_{K - \lfloor \rho K \rfloor}(d)
\label{eq:hybrid-selection}
\end{equation}

where \( \text{Bottom}_k(d) \) selects the \( k \) samples with the smallest distances, and \( \text{Top}_k(d) \) selects the \( k \) samples with the largest distances.

\subsection{Method Summary}

The total training loss is composed in the following way:

\begin{equation}
\mathcal{L}_{\text{total}} = \mathcal{L}_{\text{CE}} + \alpha \mathcal{L}_{\text{distill}} + \beta \mathcal{L}_{\text{att}} + \gamma \mathcal{L}_{\text{rep}},
\end{equation}

where $\mathcal{L}_{\text{CE}}$ is the cross-entropy loss for mixed current + replay mini-batch, and $\alpha$, $\beta$, and $\gamma$ are hyperparameters used to balance the contributions of the distillation, attraction, and repulsion losses, respectively. This balanced objective allows ProDER to effectively learn new tasks while maintaining performance on previously learned tasks.

In conclusion, unlike vanilla DER, which stores logits and selects samples randomly, or DER++, which adds distillation without semantic awareness, ProDER introduces three key enhancements:
\begin{itemize}
    \item \textbf{Prototype-based supervision} aligns feature representations semantically across tasks.
    \item \textbf{Repulsion regularization} enforces inter-class discrimination in the learned feature space.
    \item \textbf{Hybrid memory pruning} retains samples that are both central and diverse, improving the usefulness of the limited replay buffer.
\end{itemize}

\subsection{Motivation}

\noindent \textbf{Why distillation is effective.} For any two classes $c$ and $j$, the difference between their log-probabilities under the model is
\begin{equation}
\log p_\theta(c \mid x) - \log p_\theta(j \mid x)
= \frac{\ell_c(x) - \ell_j(x)}{T},
\end{equation}
where $p_\theta(c \mid x)$ denotes the predicted probability of class $c$ for input $x$,
$\ell_c(x)$ is the corresponding logit,
and $T>0$ is the temperature parameter controlling the sharpness of the distribution.
Matching the full soft probability distribution, therefore, compels a student model to reproduce
all pairwise logit differences produced by the teacher for a given input $x$, rather than merely
the top-scoring class. This enforces agreement with the teacher’s fine-grained ordering of classes.

\noindent\textbf{Geometric Interpretation.}
We next provide a conditional interpretation of the relationship between classifier logits and prototype distances. Let $f=f_\theta(x)\in\mathbb{R}^D$ denote the representation provided to the linear classifier, with
\[
\ell_c(x)=\mathbf{w}_c^\top f+b_c.
\]
Let $\boldsymbol{\mu}_c$ denote the class-mean representation of class $c$ in this feature space. The relevance of class-mean geometry and its relationship with the final linear classifier has also been observed in the neural-collapse literature, particularly during the terminal phase of training~\cite{papyan2020prevalence}. Here, however, we do not assume that classifier weights coincide with class means. In ProDER, the prototype $\mathbf{p}_c$ used by the attraction loss is the empirical class mean computed from the samples of class $c$ present in the current mixed mini-batch and can therefore be viewed as a mini-batch estimate of the corresponding class-mean representation $\boldsymbol{\mu}_c$.

For any fixed shared scale $\lambda>0$, define
\begin{equation}
\mathbf{r}_c=\mathbf{w}_c-\lambda\boldsymbol{\mu}_c,
\end{equation}
so that
\begin{equation}
\mathbf{w}_c=\lambda\boldsymbol{\mu}_c+\mathbf{r}_c,
\label{eq:w_decomp}
\end{equation}
where $\mathbf{r}_c$ denotes the residual associated with this decomposition.

\begin{proposition}[Relation between logit differences and prototype distances]
\label{prop:geometry}
For any input $x$ and any two classes $c$ and $j$,
\begin{equation}
\ell_c(x)-\ell_j(x)
=
\frac{\lambda}{2}
\Big(
\|f-\boldsymbol{\mu}_j\|^2
-
\|f-\boldsymbol{\mu}_c\|^2
\Big)
+
\delta_{cj}(x),
\label{eq:logit_diff}
\end{equation}
where
\begin{equation}
\delta_{cj}(x)
=
\underbrace{
\frac{\lambda}{2}
\left(
\|\boldsymbol{\mu}_c\|^2-
\|\boldsymbol{\mu}_j\|^2
\right)
+b_c-b_j
}_{\bar{\delta}_{cj}\text{, independent of }x}
+
(\mathbf{r}_c-\mathbf{r}_j)^\top f.
\label{eq:delta}
\end{equation}

Consequently:
\begin{enumerate}
\item[(i)]
If $\mathbf{r}_c=\mathbf{0}$ for all $c$ and
$\frac{\lambda}{2}\|\boldsymbol{\mu}_c\|^2+b_c$
is constant across classes, then $\delta_{cj}(x)=0$ and
\begin{equation}
\mathrm{softmax}
\left(
\frac{\ell(x)}{T}
\right)_c
=
\frac{
\exp\left(
-\lambda\|f-\boldsymbol{\mu}_c\|^2/(2T)
\right)
}{
\sum_k
\exp\left(
-\lambda\|f-\boldsymbol{\mu}_k\|^2/(2T)
\right)
}.
\label{eq:softmax_proto}
\end{equation}
Thus, under these conditions, the soft output is exactly determined by the squared distances from $f$ to the class-mean representations.

\item[(ii)]
In the general case,
\begin{equation}
\left|
\delta_{cj}(x)-\bar{\delta}_{cj}
\right|
\leq
\|\mathbf{r}_c-\mathbf{r}_j\|\,\|f\|,
\end{equation}
and therefore
\begin{equation}
|\delta_{cj}(x)|
\leq
|\bar{\delta}_{cj}|
+
\|\mathbf{r}_c-\mathbf{r}_j\|\,\|f\|.
\end{equation}
Hence, the deviation from the prototype-distance term consists of a class-pair-dependent constant offset and an input-dependent component controlled by the residual difference.
\end{enumerate}
\end{proposition}

\begin{proof}
Expanding the squared Euclidean distance gives
\[
\|f-\boldsymbol{\mu}_c\|^2
=
\|f\|^2
-
2\boldsymbol{\mu}_c^\top f
+
\|\boldsymbol{\mu}_c\|^2,
\]
and therefore
\[
\boldsymbol{\mu}_c^\top f
=
\frac{1}{2}
\left(
\|f\|^2
+
\|\boldsymbol{\mu}_c\|^2
-
\|f-\boldsymbol{\mu}_c\|^2
\right).
\]
Substituting~\eqref{eq:w_decomp} into
\[
\ell_c(x)-\ell_j(x)
=
(\mathbf{w}_c-\mathbf{w}_j)^\top f+b_c-b_j
\]
and applying the above identity causes the common $\|f\|^2$ terms to cancel, yielding~\eqref{eq:logit_diff} and~\eqref{eq:delta}.

Under the conditions of (i),
\[
\ell_c(x)
=
-\frac{\lambda}{2}
\|f-\boldsymbol{\mu}_c\|^2
+
\kappa(x),
\]
where $\kappa(x)$ is independent of the class index. Since softmax is invariant to adding the same quantity to every logit,~\eqref{eq:softmax_proto} follows. The first bound in (ii) follows directly from the Cauchy--Schwarz inequality, and the second follows from the triangle inequality.
\end{proof}

The proposition should be interpreted as a conditional relationship and not as an assumption that classifier weights are equal to class prototypes. It exactly separates the contribution associated with prototype distances from a class-dependent offset and a residual term. If there exists a shared scale $\lambda>0$ for which these additional terms are small for the relevant class pairs and exemplars, relative logits are approximately related to differences in squared distances to the class-mean representations.

Knowledge distillation acts on these relative outputs. For a fixed temperature and the same set of distilled classes, equality of the teacher and student soft distributions implies equality of their pairwise logit differences, up to the common additive invariance of softmax. Therefore, when the approximate conditions described above hold, preserving relative logits can also help preserve teacher output relationships that reflect the prototype-distance structure of previously observed examples.

This interpretation does not require an exact equivalence between classifier weights and prototypes. Rather, it motivates prototype regularization and logit distillation as complementary mechanisms: the former directly structures the
feature space through within-class compactness and inter-class separation, whereas the latter preserves previously learned output relationships across tasks.

\noindent\textbf{Prototype-aware Selection Strategy.}
The memory-selection strategy is designed to preserve complementary regions of each class distribution. Samples located close to the class prototype are treated as central or representative examples, since they capture features near the empirical center of the class. Samples with larger distances from the prototype are treated as peripheral or diverse examples (potentially near the decision boundaries), since they capture variations that are less represented by the central samples. These peripheral samples may include examples that are relevant to class boundaries.

By retaining both representative and peripheral examples, the replay buffer preserves information about both the central tendency and the diversity of each class. This provides a broader coverage of the previously observed feature distribution than selecting only samples closest to the prototype.

\noindent\textbf{Why it works.}
Prototype-aware exemplar selection and semantic prototype regularization are naturally connected through the learned feature space. The attraction and repulsion losses shape this representation, while exemplar selection uses distances to the resulting class prototypes to identify central and peripheral samples. The selected examples are then replayed together with their stored logits, allowing knowledge distillation to preserve output-level information associated with these samples.

This creates a feedback loop between selection and representation, so that the stored exemplars remain informative as the feature clusters evolve.
The loop works as follows. Prototypes determine which exemplars are stored, namely core samples that anchor
the cluster center and peripheral samples that capture additional variation within the class. The stored exemplars retain soft logits that preserve output-level
relationships learned when the exemplars were stored. As discussed in Proposition~\ref{prop:geometry}, when the additional residual and class-dependent terms are small, these relative outputs can approximately reflect differences in squared distances to the class-mean representations. Knowledge distillation encourages the student to preserve these previously learned output relationships during subsequent tasks, complementing the feature-space structure promoted by prototype regularization.

\begin{figure}[htbp]
  \centering
  \includegraphics[width=0.8\textwidth]{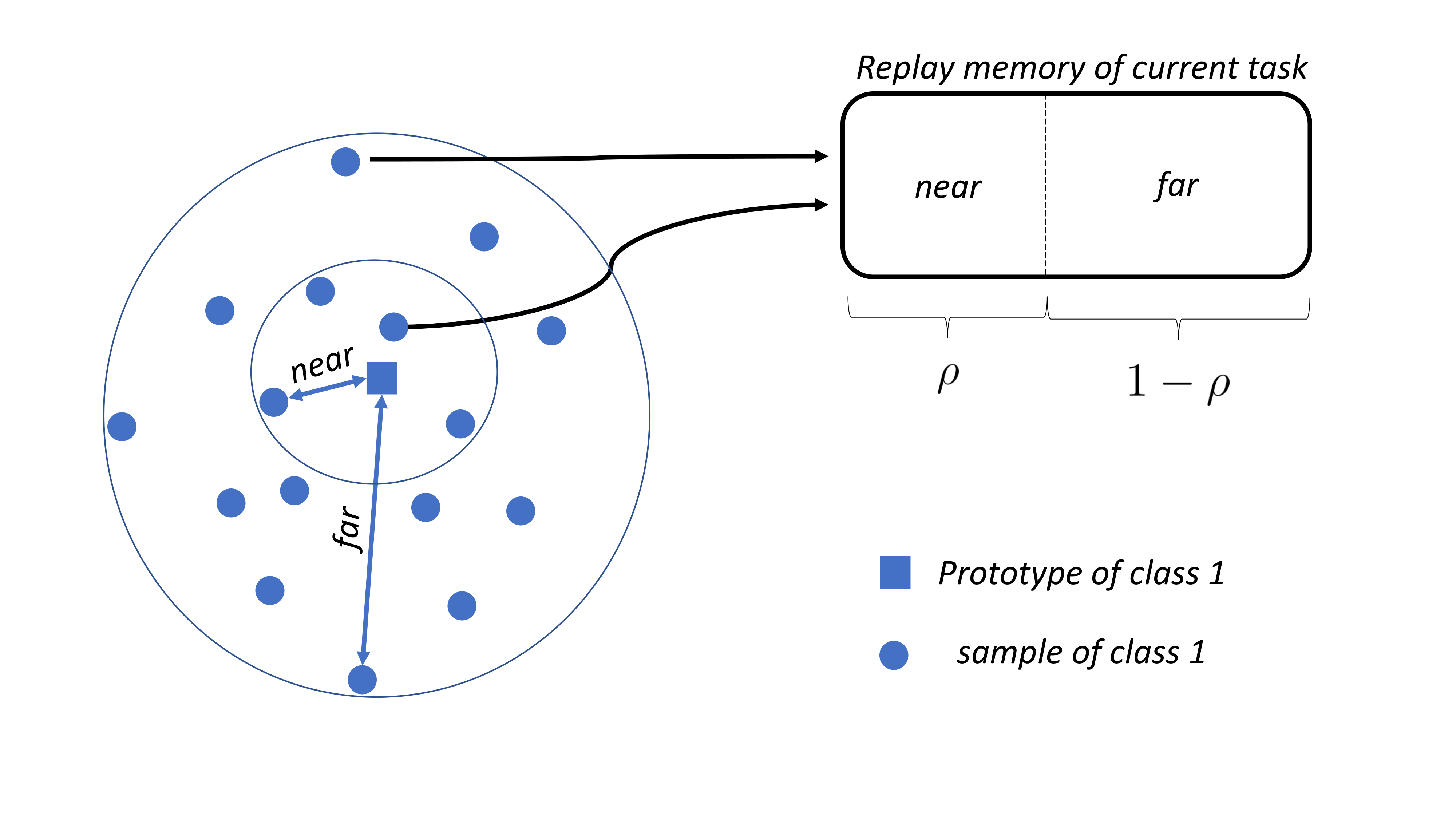}
  \caption{\textbf{Prototype-aware selection strategy to populate the replay memory}. Instead of the random selection strategy used by DER, we propose a novel sample selection method that selects both samples close to the class prototype and those that lie farther away, potentially near the decision boundaries. This approach enables the model to focus on preserving the semantic core of each class while also maintaining awareness of inter-class boundaries, improving both representation stability and discrimination. $\rho$ controls the proportion of near samples and far samples.}
  \label{fig:prototype_aware_selection_strategy}
\end{figure}

\subsection{Complexity Analysis}


{We denote by $N$ the number of samples in the current mixed mini-batch, $C_t$ the total number of classes observed up to task $t$, $D$ the feature dimension, $M$ the total replay-memory capacity, $R$ the number of replay samples drawn per mini-batch, and $K_b$ the number of classes represented in the current mixed mini-batch. By definition, $K_b \leq C_t$.

\subsubsection{Replay Memory Management}
The system maintains a fixed-size replay buffer with total capacity \(M\). Assuming an approximately balanced allocation among the classes observed up to task \(t\), each class retains approximately

\[
N_c \approx \frac{M}{C_t}
\]

samples, where \(C_t\) denotes the number of classes observed up to task \(t\).

At each task transition, we apply \emph{hybrid pruning} to maintain the bounded per-class memory allocation. For each class \(c\) with \(N_c\) stored examples, the distance between each feature representation and the corresponding class prototype is computed. The samples are then sorted according to these distances, and the closest and farthest examples are selected according to the prototype-aware memory strategy.

Computing the distances for class \(c\) requires \(O(N_cD)\) operations, while sorting them requires \(O(N_c\log N_c)\). Therefore, the pruning cost for class \(c\) is

\[
O(N_cD + N_c\log N_c).
\]

Repeating this process over the \(C_t\) observed classes, and using

\[
\sum_{c=1}^{C_t} N_c \leq M,
\]

the total memory-update cost is bounded by

\begin{equation}
O\!\left(
\sum_{c=1}^{C_t}
\left(
N_cD + N_c\log N_c
\right)
\right)
\leq
O\!\left(
MD + M\log M
\right).
\label{eq:memory-selection-complexity}
\end{equation}

Therefore, updating the replay memory at the end of each task has complexity

\[
O(MD + M\log M).
\]

\subsubsection{Loss}
\label{subsub:loss_computation}

The computational cost of the training objective can be divided into
three main parts:
1) cross-entropy (CE) classification loss,
2) knowledge distillation (KD), and
3) semantic prototype regularization.

\noindent \textbf{Knowledge distillation (soft-target) loss}\\

For replayed samples, we store the logits produced by the model when
the samples are added to memory. During replay, these stored logits are
used as soft targets through a temperature-scaled KL-divergence loss.
For each replay sample, computing the softmax, log-softmax, and
KL-divergence requires $O(C_t)$ operations, where $C_t$ denotes the
number of classifier outputs corresponding to the classes observed up
to task $t$. Therefore, for $R$ replay samples, the cost is

\begin{equation}
O(RC_t).
\end{equation}

per mini-batch.

\noindent \textbf{Prototype attraction and repulsion losses}\\

The class prototypes are computed directly from the feature
representations in the current mixed mini-batch. Let $N$ be the number
of samples in the mini-batch, $D$ the feature dimension, and $K_b$ the
number of classes represented in that mini-batch.

\begin{itemize}

    \item \textbf{Prototype computation:}
    {
    The prototype of each represented class is obtained by averaging
    the feature vectors of the samples belonging to that class. This
    requires one pass over the feature representations and has a cost
    of $O(ND)$.}

    \item \textbf{Prototype attraction:}
    
    The attraction term measures the squared distance between each
    feature vector and the prototype of its corresponding class.
    Computing these distances for all $N$ samples also requires
    $O(ND)$ operations.

    \item \textbf{Prototype repulsion:}
    
    A sample-level repulsion strategy would compare samples belonging
    to different classes with one another. In the worst case, this
    requires $O(N^2D)$ operations, since the number of pairwise
    comparisons grows quadratically with the mini-batch size.

    In ProDER, repulsion is instead applied between class prototypes.
    The $K_b$ prototypes are first $\ell_2$-normalized, which requires
    $O(K_bD)$ operations. Pairwise Euclidean distances are then computed
    between the normalized prototypes. This step has a cost of

    \begin{equation}
    O(K_b^2D).
    \end{equation}

    The margin-based penalty is evaluated only over the unique
    prototype pairs, adding an $O(K_b^2)$ term. Since the pairwise
    distance computation is the dominant operation, the overall cost
    of the repulsion component remains

    \begin{equation}
    O(K_b^2D).
    \end{equation}

    In this way, the repulsion term works with class-level
    representations rather than individual negative sample pairs.
    The pairwise cost is therefore reduced from $O(N^2D)$ to
    $O(K_b^2D)$. This is useful when the number of classes represented
    in the mini-batch is much smaller than the number of samples.

\end{itemize}

Combining prototype computation, attraction, and repulsion, the total
cost of the semantic prototype regularization for one mixed mini-batch
is

\begin{equation}
O\!\left(ND + K_b^2D\right).
\label{eq:proder-complexity}
\end{equation}

When the number of represented classes is small enough that
$K_b^2 \ll N$, the $ND$ term dominates, and the complexity can be
approximated as

\begin{equation}
O\!\left(ND + K_b^2D\right)
\approx
O(ND).
\label{eq:proder-complexity-simplified}
\end{equation}

Thus, the prototype computation and attraction terms scale linearly
with the number of samples in the mini-batch, while the quadratic term
depends only on the number of classes represented in that mini-batch.
Compared with a sample-wise pairwise repulsion term, which can require
$O(N^2D)$ operations, the prototype-based formulation limits the
additional pairwise computation to $O(K_b^2D)$.

\subsubsection{Per-batch combined cost and per-task training}

A single training step (mini-batch) processes a model's forward/backward
passes, as well as all auxiliary losses. Approximate costs:

\begin{itemize}

  \item Cross-entropy (labels) loss:
  
  \(O(N\cdot C_t)\) for computing logits and loss terms across
  \(C_t\) outputs on a mini-batch of \(N\) samples.

  \item Knowledge Distillation:
  
  \(O(R\cdot C_t) \overset{R < N}{=} O(NC_t)\).

  \item Prototype computation: \(O(ND)\).

  \item Prototype attraction: \(O(ND)\).

  \item Prototype repulsion:
  
  \(O(K_b^2D)\), where \(K_b\) denotes the number of classes
  represented in the current mixed mini-batch. The $\ell_2$
  normalization of the \(K_b\) prototypes adds \(O(K_bD)\), which
  does not change the dominant \(O(K_b^2D)\) cost of the pairwise
  prototype distances.

\end{itemize}

Combining these, the cost to compute the loss function for a mini-batch is

\begin{equation}
O\!\left(NC_t + NC_t + ND + K_b^2D\right)
\;\overset{C_t \ll D}{=}\;
O\!\left(ND + K_b^2D\right)
\label{eq:proder-total-complexity}
\end{equation}

The prototype-based repulsion therefore retains the same asymptotic
complexity as the previous prototype-level formulation. Its advantage
over a sample-wise pairwise repulsion strategy is that the pairwise
term depends on the number of classes represented in the current
mini-batch rather than on the number of samples, reducing the
corresponding cost from \(O(N^2D)\) to \(O(K_b^2D)\).

Moreover, at the end of each task, updating the memory has a cost of

\(O(MD + M\log M)\), where \(M\) denotes the total replay-memory
capacity.

\section{Experimental Setting}
\label{sec:ExpSetting}

In this section, we provide all the specific details concerning the implementations of our work.
Specifically, we provide all the hyperparameters used in the model and training configuration.
Similarly, we provide for each CL approach the hyperparameters used.
Eventually, we provide the metrics used to evaluate the results obtained in the several tested scenarios.

\subsection{Fault Diagnosis Model Implementation}

For all experiments in this study, we use a unified static architecture designed for temporal fault diagnosis from sequential sensor data. This model remains fixed across all scenarios, including static evaluation and CL settings, ensuring that performance differences arise solely from the learning strategies rather than architectural changes.

\textbf{Model Architecture:} The core model is a recurrent neural network consisting of:
\begin{itemize}
    \item A single bidirectional Gated Recurrent Unit (GRU) layer with $150$ hidden units in each direction, resulting in a $300$-dimensional output per timestep.
    \item A dropout layer with a dropout probability of $0.3$, applied to the output of the GRU to prevent overfitting.
    \item A fully connected classification layer mapping the temporal representation to class logits. This layer is expanded dynamically when new classes are introduced in CL, without reinitializing existing parameters.
\end{itemize}

\subsection{Training Configuration} 
The model is trained using the Adam optimizer with a fixed learning rate of $0.001$ and a batch size of $4$. All training is conducted for $50$ epochs per task or dataset partition, without early stopping or learning rate scheduling. The input sequences are processed in batch-first format, and

all experiments are repeated using five random seeds. The same set of
seeds is used across the evaluated methods to provide a consistent
comparison.
 This model forms the backbone for all subsequent experiments and serves as a standard temporal encoder for sensor-based fault diagnosis tasks. All modifications in CL settings are applied on top of this fixed architecture and training setup.

\subsection{CL Approaches} 
For the hyperparameters of the CL approaches, a replay ratio $0.5$ is used per batch, and the memory buffer with a size of 363 (like the case of ER and DER++) is updated at the end of each task using the hybrid selection mechanism. Prototypes are recomputed on-the-fly during training to reflect evolving feature dynamics.

For the reported experiments, the ProDER hyperparameters were selected separately for each scenario. We use $(\alpha,\beta,\gamma)=(4,8.2,0.5)$ for Scenario~1, $(4,10,0.5)$ for Scenario~2, $(1,10,2.0)$ for Scenario~3, and $(2,7,2)$ for Scenario~4. The prototype-selection ratio $\rho$, which controls the balance between samples close to and far from the class prototype, is set to $0.50$, $0.82$, $0.62$, and $0.75$ for Scenarios~1--4, respectively. The repulsion margin $m$ is also selected per scenario and is set to $1.4$ for Scenarios~1 and $1.2$ for Scenarios~2 , 3 and 4.

To select the ProDER-specific hyperparameters under a fixed computational budget, we performed a constrained search over the joint space of $\beta$, $\gamma$, $\rho$, and $m$ using the Golden search optimization procedure, with a maximum of 100 search iterations. The same tuning budget was used for the other compared methods to maintain a fair experimental setting. Importantly, $\alpha$ was not independently tuned for ProDER. Instead, it was kept equal to the changed DER++ distillation coefficient in each scenario. This keeps the distillation strength consistent with DER++ while allowing the evaluation to focus on the additional contribution of the prototype-based losses and prototype-aware memory construction.

\subsection{Evaluation}
\label{Eval}
In this section, we first present the dataset used to evaluate our proposed approaches. We then assess the performance of the various methods described in Section~\ref{CL_method} across different CL scenarios. Finally, we conclude with a discussion of the results and their implications. 
We evaluate the performance of the continual learning (CL) methods using final accuracy (\textbf{ACC}), performance gap (\textbf{gap})~\cite{FANAN2025110459}, weighted precision (\textbf{W-Prec.}), weighted recall (\textbf{W-Rec.}), weighted F1-score (\textbf{W-F1.}), and macro F1-score. These metrics allow us to assess not only the final classification accuracy, but also the stability of each method across  classes and previously learned tasks.

Let $\text{acc}_{t,j}$ denote the accuracy on task $j$ after the model has been trained up to task $t$. For a test set $\mathcal{D}^{\mathrm{test}}_j$ with $N_j$ samples, it is defined as:
\begin{equation}
\text{acc}_{t,j} =
\frac{1}{N_j}
\sum_{n=1}^{N_j}
\mathbf{1}
\left\{
f_{\theta_t}\left(\mathbf{x}_{j,n}\right)
=
y_{j,n}
\right\}.
\label{eq:task-accuracy}
\end{equation}

The final accuracy is then computed after training on the last task $T$ as the average accuracy over all tasks:
\begin{equation}
\mathrm{ACC}
=
\frac{1}{T}
\sum_{j=1}^{T}
\text{acc}_{T,j}.
\label{eq:final-accuracy}
\end{equation}

In addition to accuracy, we report the \textbf{gap}, which measures the difference between the ideal upper-bound performance obtained by \textit{Joint Training}, where all tasks are trained together, and the performance achieved by each CL method. The gap is defined as:
\begin{equation}
\mathrm{gap}
=
\mathrm{ACC}_{\mathrm{Joint}}
-
\mathrm{ACC}_{\mathrm{CL}},
\label{eq:performance-gap}
\end{equation}
where $\mathrm{ACC}_{\mathrm{Joint}}$ is the final accuracy of the jointly trained model, and $\mathrm{ACC}_{\mathrm{CL}}$ is the final accuracy of the continual learning method. A smaller gap indicates that the CL method is closer to the joint-training upper bound and loses less performance due to sequential learning.

Because the fault classes are not always equally represented, we also report weighted precision, weighted recall, weighted F1-score, and macro F1-score. For each class $c$, precision, recall, and F1-score are computed as:
\begin{equation}
\mathrm{Precision}_c =
\frac{\mathrm{TP}_c}{\mathrm{TP}_c + \mathrm{FP}_c},
\label{eq:class-precision}
\end{equation}

\begin{equation}
\mathrm{Recall}_c =
\frac{\mathrm{TP}_c}{\mathrm{TP}_c + \mathrm{FN}_c},
\label{eq:class-recall}
\end{equation}

\begin{equation}
\mathrm{F1}_c =
\frac{
2 \cdot \mathrm{Precision}_c \cdot \mathrm{Recall}_c
}{
\mathrm{Precision}_c + \mathrm{Recall}_c
}.
\label{eq:class-f1}
\end{equation}

The weighted F1-score is computed by weighting each class F1-score according to the number of test samples in that class:
\begin{equation}
\mathrm{Weighted\ F1}
=
\sum_{c=1}^{C}
\frac{N_c}{N}
\mathrm{F1}_c,
\label{eq:weighted-f1}
\end{equation}
where $N_c$ is the number of test samples belonging to class $c$, $N$ is the total number of test samples, and $C$ is the number of classes. Weighted precision and weighted recall are computed in the same way. Finally, macro F1-score gives equal importance to all classes and is defined as:
\begin{equation}
\mathrm{Macro\ F1}
=
\frac{1}{C}
\sum_{c=1}^{C}
\mathrm{F1}_c.
\label{eq:macro-f1}
\end{equation}

Accuracy and gap measure the overall retention of knowledge, while weighted and macro scores show whether the model maintains balanced performance across both frequent and less frequent fault classes. For consistency of presentation, all classification performance metrics reported in the tables and figures are expressed on a percentage scale (0--100\%), obtained by multiplying their corresponding values on the $[0,1]$ scale by 100.

\subsection{Dataset}
 Each fault type was simulated using 22 different resistance values. Faults were introduced at \( t = 0.01 \) seconds and cleared at \( t = 0.02 \) seconds, defining a fault-active interval of \( t_f = [0.01 - 0.02] \) seconds. The interval prior to fault activation, \( t_h = [0 - 0.01] \) seconds, remained fault-free~\cite{mainpaper}.  As a preprocessing step, we apply standard score normalization (Z-score) to scale all relevant features in the dataset before training our GRU-based model (

z-score normalization parameters are estimated from the training partition and then applied to the corresponding held-out data.
)
. After normalization, the data is segmented into overlapping time windows to prepare it for sequential modeling using a GRU network. The dataset is sorted by fault class, and each class is trimmed to ensure the number of samples is divisible by the predefined window size, set to 12. To create temporal sequences, a sliding window approach is used with a step size equal to half the window length (i.e., 6).

To prevent information leakage between overlapping sequences, the data are split into training and held-out partitions before the sliding-window procedure is applied. Consequently, windows generated from the same simulated event remain within the same partition and no overlapping observations are shared between the training and evaluation sets.

\section{Results}
\label{sec:results}

In this section, we present the evaluation results for each CL scenario described in Section~\ref{sc}, using the methods introduced in Section~\ref{CL_method}. Table~\ref{Tab:tab_results} provides an overview of the main results, reporting accuracy and gap to Joint Training for all four scenarios together with their average; the complete set of metrics for each scenario is reported in the corresponding subsection.

\subsection{Evaluation of Scenario 1 -- Fault Type Classification with Two New Fault Classes}

Scenario 1 contains five consecutive tasks and represents the basic class-incremental setting for fault type classification. We use this scenario as a starting point because it follows the common CL setup, where new fault classes are introduced over time. The results are shown in Table~\ref{tab:scenario1_results} and Figures~\ref{fig1_1} and~\ref{fig:1_2}.
Fine-Tuning shows strong forgetting after the final task, with an average accuracy of only $15.4 \pm 1.0$ and a macro-F1 score of $5.1 \pm 0.0$. EWC does not improve this result, reaching an accuracy of $14.9 \pm 0.0$, while LwF performs slightly better at $16.5 \pm 1.0$. Their low macro-F1 scores show that regularization alone is not sufficient to preserve performance across the previously learned fault classes in this scenario.

Replay-based methods perform considerably better. ER reaches an accuracy of $32.8 \pm 1.0$, while DER++ improves this to $40.1 \pm 1.0$ through replay and output-level knowledge preservation. DPDMR achieves a similar but slightly higher accuracy of $40.9 \pm 2.0$, with a macro-F1 score of $40.3 \pm 2.0$. ProDER gives the best result among the CL methods, reaching $46.1 \pm 2.0$ accuracy and reducing the gap from Joint Training to $5.8 \pm 1.0$. Compared with DER++, this corresponds to an improvement of about 6.0 in average accuracy. The same trend can be seen in the other metrics, where ProDER obtains a weighted F1 and macro-F1 of $45.8 \pm 2.0$ and the highest weighted precision among the CL methods at $53.2 \pm 2.0$.

ProDER achieves the highest mean accuracy among the evaluated CL methods in Scenario~1, while Joint Training remains the reference performance at $52.0 \pm 1.0$.

\begin{figure}[H]
  \centering
  \begin{subfigure}[b]{0.49\textwidth}
    \includegraphics[width=\textwidth]{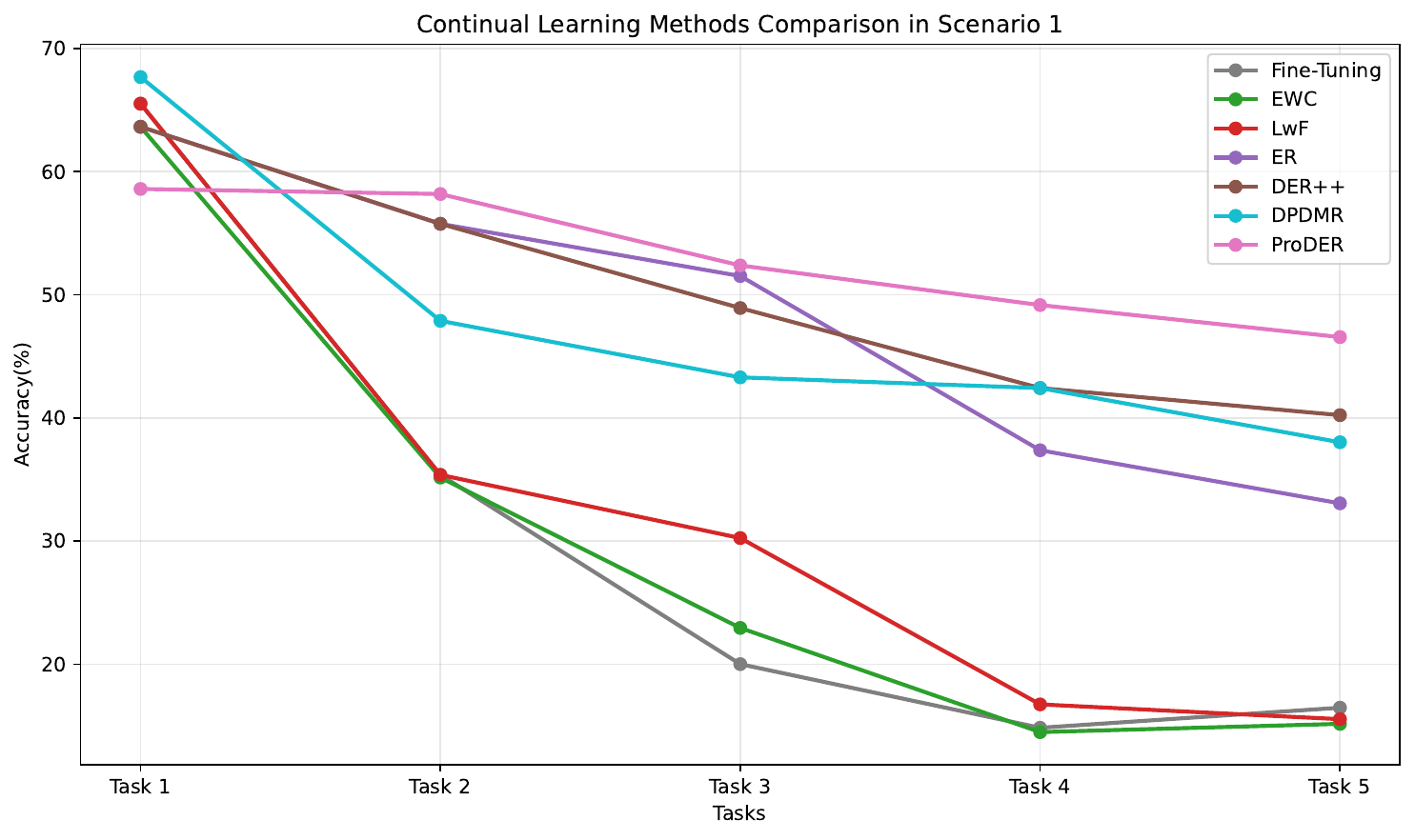}
    \caption{Test accuracy of each method and baselines}
    \label{fig1_1}
  \end{subfigure}
  \hfill
  \begin{subfigure}[b]{0.49\textwidth}
    \includegraphics[width=\textwidth]{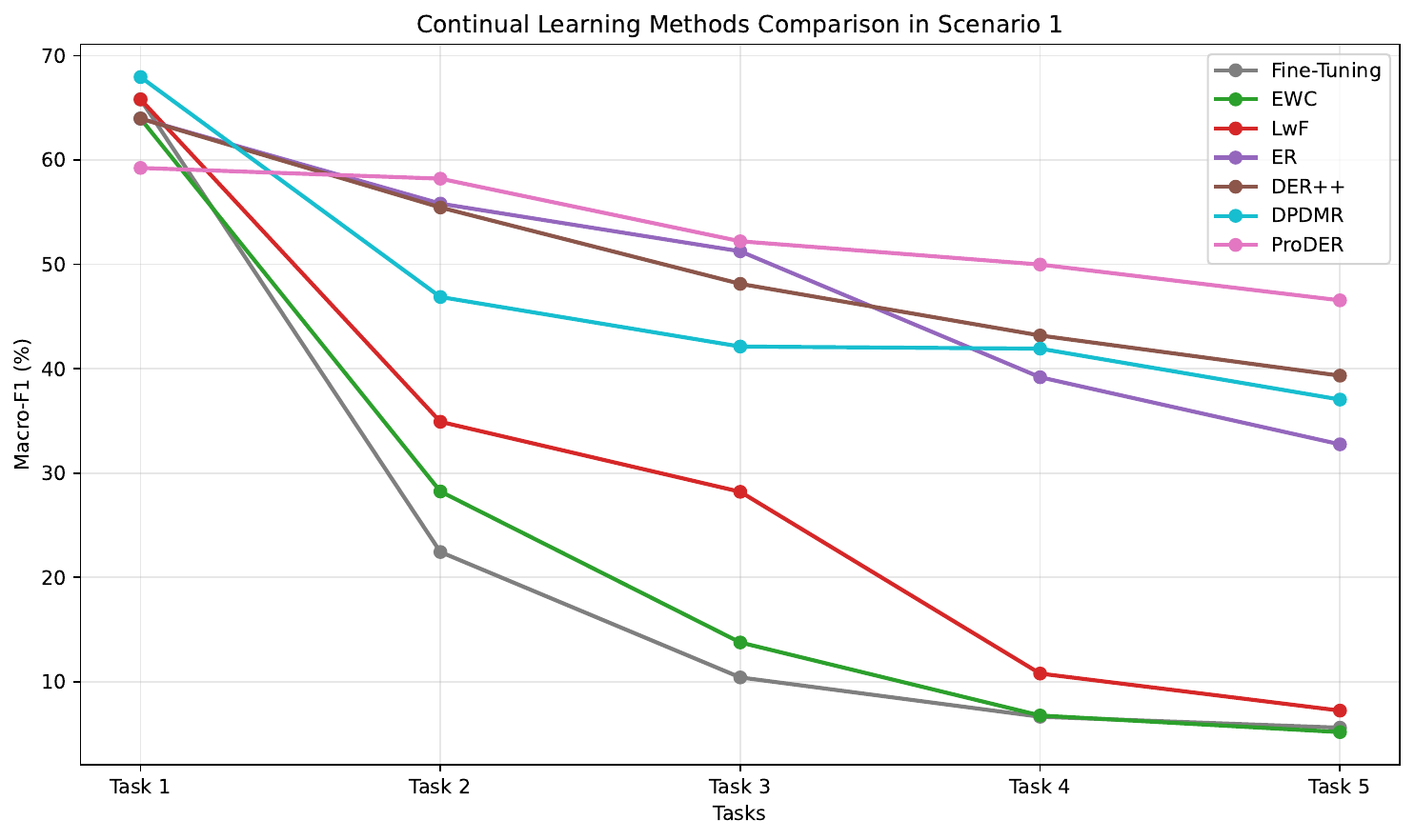}
    \caption{Test Macro-F1 of each method and baselines}
    \label{fig:1_2}
  \end{subfigure}
  \caption{Task-wise accuracy and Macro-F1 for one fixed seed in Scenario 1.}
  \label{fig1}
\end{figure}

\begin{table*}[t]
\centering
\caption{Evaluation of CL methods in Scenario 1 using  accuracy, gap, and class-balanced classification metrics (Mean over 5 seeds). }
\label{tab:scenario1_results}

\scriptsize
\renewcommand{\arraystretch}{1.15}
\setlength{\tabcolsep}{3pt}

\resizebox{\textwidth}{!}{%
\begin{tabular}{lcccccc}
\hline
\textbf{Method} &
\textbf{ACC} &
\textbf{Gap} &
\textbf{W-F1} &
\textbf{W-Prec.} &
\textbf{W-Rec.} &
\textbf{Macro-F1}
\tabularnewline
\hline

Joint Training
& $52.0 \pm 1.0$
& $0.0 \pm 0.0$
& $52.4 \pm 1.0$
& $54.2 \pm 1.0$
& $52.0 \pm 1.0$
& $52.4 \pm 1.0$
\tabularnewline

Fine-Tuning
& $15.4 \pm 1.0$
& $36.5 \pm 1.0$
& $5.1 \pm 0.0$
& $8.2 \pm 5.0$
& $15.4 \pm 1.0$
& $5.1 \pm 0.0$
\tabularnewline

EWC
& $14.9 \pm 0.0$
& $37.1 \pm 1.0$
& $5.4 \pm 1.0$
& $10.6 \pm 5.0$
& $14.9 \pm 0.0$
& $5.4 \pm 1.0$
\tabularnewline

LwF
& $16.5 \pm 1.0$
& $35.5 \pm 2.0$
& $7.7 \pm 1.0$
& $10.7 \pm 5.0$
& $16.5 \pm 1.0$
& $7.7 \pm 1.0$
\tabularnewline

ER
& $32.8 \pm 1.0$
& $19.2 \pm 1.0$
& $32.5 \pm 2.0$
& $43.1 \pm 3.0$
& $32.8 \pm 1.0$
& $32.5 \pm 2.0$
\tabularnewline

DER++
& $40.1 \pm 1.0$
& $11.9 \pm 2.0$
& $39.2 \pm 2.0$
& $47.3 \pm 3.0$
& $40.1 \pm 1.0$
& $39.2 \pm 2.0$
\tabularnewline

DPDMR
& $40.9 \pm 2.0$
& $11.0 \pm 3.0$
& $40.3 \pm 2.0$
& $51.9 \pm 2.0$
& $40.9 \pm 2.0$
& $40.3 \pm 2.0$
\tabularnewline

ProDER
& \textbf{$46.1 \pm 2.0$}
& \textbf{$5.8 \pm 1.0$}
& \textbf{$45.8 \pm 2.0$}
& \textbf{$53.2 \pm 2.0$}
& \textbf{$46.1 \pm 2.0$}
& \textbf{$45.8 \pm 2.0$}
\tabularnewline

\hline
\end{tabular}%
}

\end{table*}

\subsection{Evaluation of Scenario 2 -- Fault Type Classification with One New Fault Class}

Scenario 2 is more difficult than Scenario 1 because the model is updated over a longer sequence of tasks. Instead of introducing two new fault classes at a time, only one new fault class is added in each step resulting in nine consecutive tasks. This setting is closer to a practical deployment case, where new fault types may become available gradually and the model must be updated without waiting for several new classes to be collected together.

The results are reported in Table~\ref{tab:scenario2_results} and Figures~\ref{fig2_1} and~\ref{fig2_2}. 
Compared with Scenario 1, the performance of all CL methods decreases, showing that the longer task sequence makes forgetting more severe. Fine-Tuning reaches an average accuracy of only $9.3 \pm 0.0$ and a macro-F1 score of $2.0 \pm 1.0$. EWC slightly improves the accuracy to $9.9 \pm 1.0$, while LwF reaches $9.2 \pm 0.0$. Their low macro-F1 scores show that regularization alone is not sufficient to preserve previously learned fault classes in this one-class incremental setting.

Replay-based methods perform considerably better. ER reaches an accuracy of $32.7 \pm 3.0$, while DER++ improves this to $39.9 \pm 2.0$ and achieves a macro-F1 score of $39.9 \pm 2.0$. DPDMR obtains an accuracy of $35.9 \pm 3.0$, performing better than ER but below DER++ in this scenario. This confirms the importance of replay when the model is exposed to a longer sequence of small class updates.

ProDER achieves the strongest performance among the evaluated CL methods, reaching an accuracy of 44.2 and reducing the gap to Joint Training to 7.8. This corresponds to an improvement of approximately 4.3 over DER++. ProDER also obtains a macro-F1 score of 44.8, compared with 39.9 for DER++, indicating that the improvement is not limited to overall accuracy but is also reflected in more balanced performance across the fault classes.

\begin{figure}[H]
  \centering
  \begin{subfigure}[b]{0.49\textwidth}
    \includegraphics[width=\textwidth]{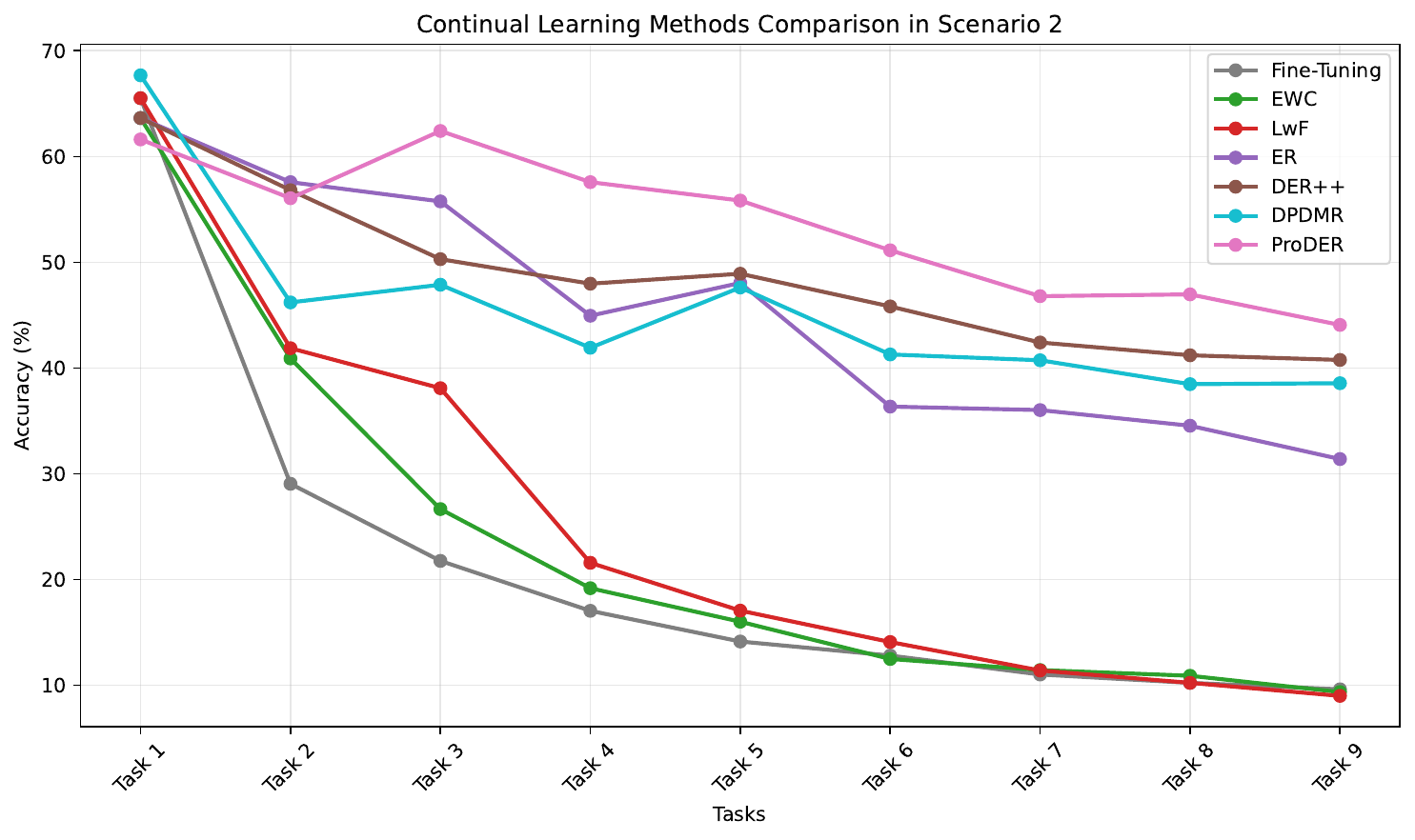}
    \caption{Test accuracy of each method and baselines}
    \label{fig2_1}
  \end{subfigure}
  \hfill
  \begin{subfigure}[b]{0.49\textwidth}
    \includegraphics[width=\textwidth]{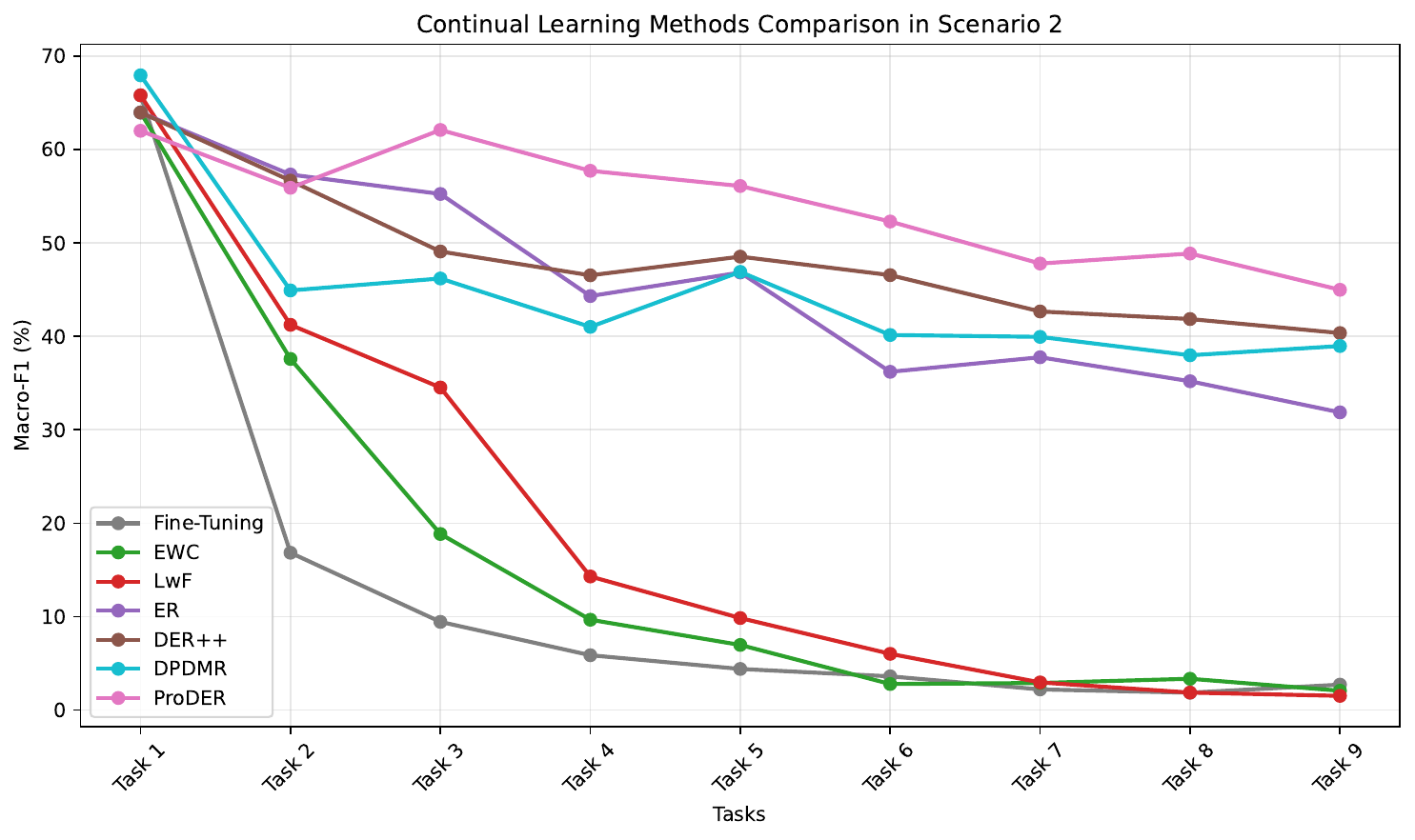}
    \caption{Test Macro-F1 of each method and baselines.}
    \label{fig2_2}
  \end{subfigure}
  \caption{Task-wise accuracy and Macro-F1 for one fixed seed in Scenario 2.}
  \label{fig:res_1}
\end{figure}

\begin{table*}[t]
\centering
\caption{Evaluation of CL methods in Scenario 2 using  accuracy, gap, and class-balanced classification metrics (Mean over 5 seeds). }
\label{tab:scenario2_results}

\scriptsize
\renewcommand{\arraystretch}{1.15}
\setlength{\tabcolsep}{3pt}

\resizebox{\textwidth}{!}{%
\begin{tabular}{lcccccc}
\hline
\textbf{Method} &
\textbf{ACC} &
\textbf{Gap} &
\textbf{W-F1} &
\textbf{W-Prec.} &
\textbf{W-Rec.} &
\textbf{Macro-F1}
\tabularnewline
\hline

Joint Training & $52.0 \pm 1.0$ & $0.0 \pm 0.0$ & $52.4 \pm 1.0$ & $54.2 \pm 1.0$ & $52.0 \pm 1.0$ & $52.4 \pm 1.0$ \tabularnewline

Fine-Tuning & $9.3 \pm 0.0$ & $42.7 \pm 1.0$ & $2.0 \pm 1.0$ & $8.0 \pm 10.0$ & $9.3 \pm 0.0$ & $2.0 \pm 1.0$ \tabularnewline

EWC & $9.9 \pm 1.0$ & $42.1 \pm 2.0$ & $3.1 \pm 1.0$ & $12.7 \pm 2.0$ & $9.9 \pm 1.0$ & $3.1 \pm 1.0$ \tabularnewline

LwF & $9.2 \pm 0.0$ & $42.8 \pm 1.0$ & $1.9 \pm 0.0$ & $5.4 \pm 5.0$ & $9.2 \pm 0.0$ & $1.9 \pm 0.0$ \tabularnewline

ER & $32.7 \pm 3.0$ & $19.3 \pm 4.0$ & $32.9 \pm 3.0$ & $46.5 \pm 3.0$ & $32.7 \pm 3.0$ & $32.9 \pm 3.0$ \tabularnewline

DER++ & $39.9 \pm 2.0$ & $12.0 \pm 2.0$ & $39.9 \pm 2.0$ & $47.8 \pm 3.0$ & $39.9 \pm 2.0$ & $39.9 \pm 2.0$ \tabularnewline

DPDMR & $35.9 \pm 3.0$ & $16.0 \pm 3.0$ & $35.8 \pm 4.0$ & $46.8 \pm 7.0$ & $35.9 \pm 3.0$ & $35.8 \pm 4.0$ \tabularnewline

ProDER & \textbf{$44.2 \pm 2.0$}
& \textbf{$7.8 \pm 2.0$}
& \textbf{$44.8 \pm 2.0$}
& \textbf{$55.3 \pm 1.0$}
& \textbf{$44.2 \pm 2.0$}
& \textbf{$44.8 \pm 2.0$}
\tabularnewline

\hline
\end{tabular}%
}

\end{table*}

\subsection{Evaluation of Scenario 3 -- Fault Type Classification with Known Faults in a New Grid Zone}

In this scenario, each task involves learning progressively more information about an already known fault type, but within a newly encountered zone. In this setting, the main challenge is not the introduction of completely new fault classes, but the change in the data distribution caused by the new operating zone. This corresponds to a domain-incremental learning problem, where the model must adapt to new zone-specific patterns while retaining what it has already learned from previous zones.

The results are reported in Table~\ref{tab:scenario3_results} and Figures~\ref{sc3_1} and~\ref{sc3_2}. 
 Compared with Scenario 2, the results improve for most CL methods. This is expected because the task sequence is shorter, with four tasks instead of nine, and the model is exposed to the same fault types across different zones. As a result, forgetting is less severe than in the one-class incremental setting. Fine-Tuning reaches an accuracy of $34.9 \pm 2.0$, while EWC and LwF obtain $31.8 \pm 4.0$ and $27.2 \pm 4.0$, respectively. Although these methods perform better than in Scenario 2, they still remain well below the replay-based approaches.

Replay again provides a clear advantage. ER reaches an accuracy of $38.4 \pm 2.0$, while DER++ improves this to $39.6 \pm 2.0$. DPDMR performs particularly well in this scenario, achieving $46.8 \pm 1.0$ accuracy and a macro-F1 score of $46.8 \pm 1.0$. This suggests that its prototype-guided replay strategy is well suited to the zone-based task sequence.

ProDER achieves the best performance among the CL methods, reaching an accuracy of $51.0 \pm 1.0$ and reducing the gap from Joint Training to only $1.0 \pm 1.0$. It also obtains a weighted F1 and macro-F1 of $51.3 \pm 1.0$. Compared with DPDMR, ProDER improves the average accuracy by about 4.2. Its performance is also substantially closer to Joint Training than the other CL baselines, showing that the combination of replay and prototype-based regularization is effective under the zone shifts considered in this scenario.

\begin{figure}[H]
  \centering
  \begin{subfigure}[b]{0.49\textwidth}
    \includegraphics[width=\textwidth]{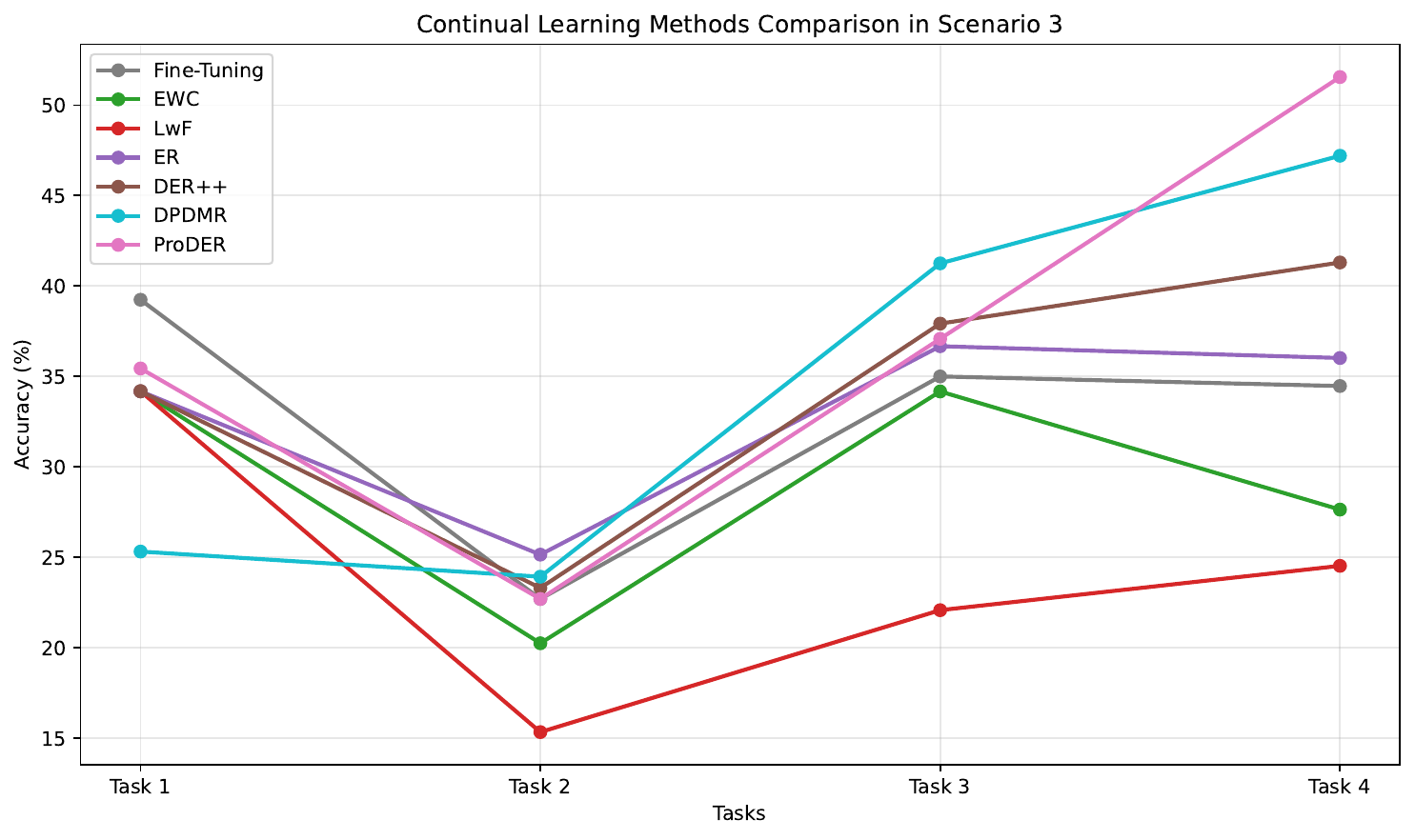}
    \caption{Test accuracy of each method and baselines}
    \label{sc3_1}
  \end{subfigure}
  \hfill
  \begin{subfigure}[b]{0.49\textwidth}
    \includegraphics[width=\textwidth]{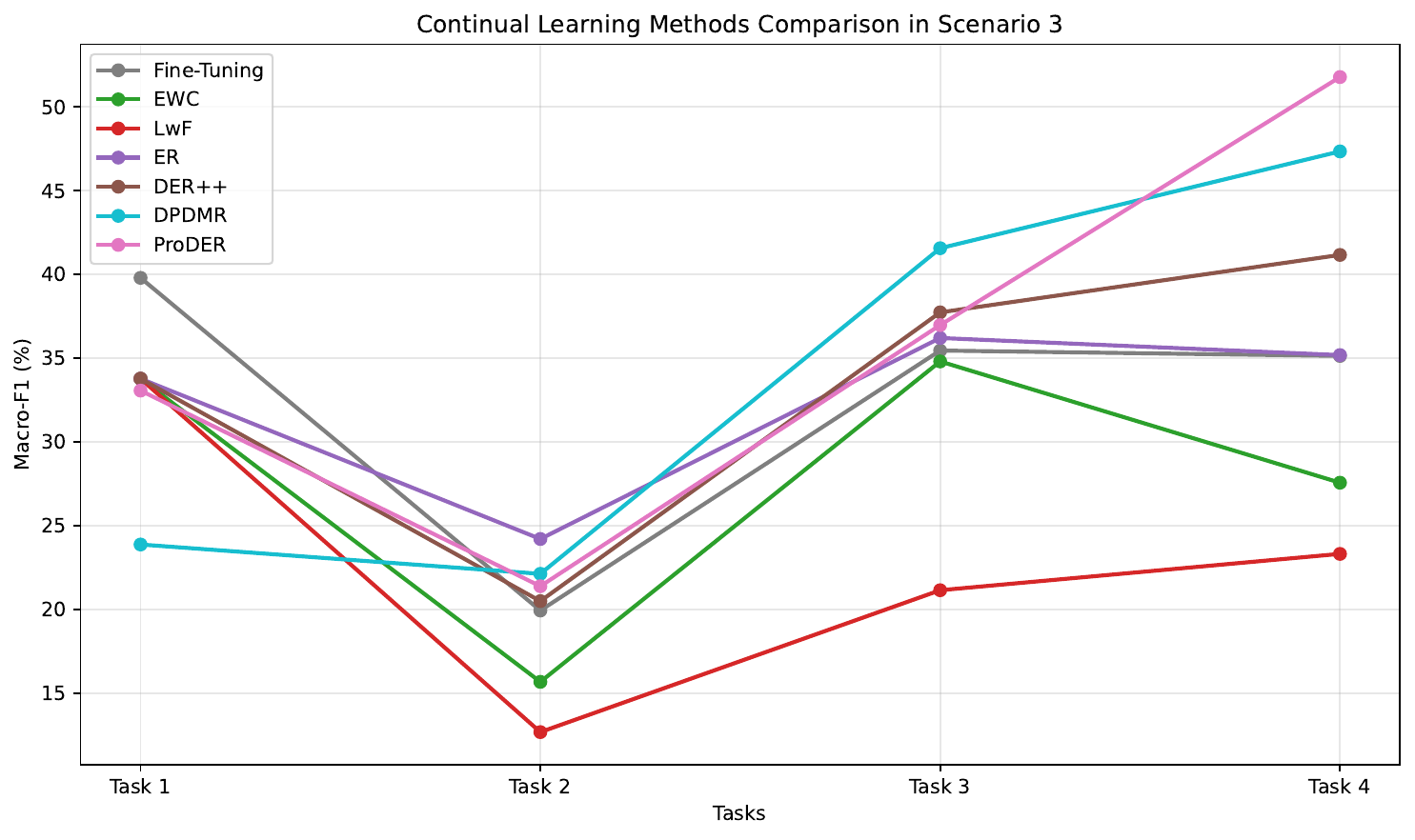}
    \caption{Test Macro-F1 of each method and baselines.}
    \label{sc3_2}
  \end{subfigure}
  \caption{Task-wise accuracy and Macro-F1 for one fixed seed in Scenario 3.}
  \label{sc3}
\end{figure}

\begin{table*}[t]
\centering
\caption{Evaluation of CL methods in Scenario 3 using  accuracy, gap, and class-balanced classification metrics (Mean over 5 seeds). }
\label{tab:scenario3_results}

\scriptsize
\renewcommand{\arraystretch}{1.15}
\setlength{\tabcolsep}{3pt}

\resizebox{\textwidth}{!}{%
\begin{tabular}{lcccccc}
\hline
\textbf{Method} &
\textbf{ACC} &
\textbf{Gap} &
\textbf{W-F1} &
\textbf{W-Prec.} &
\textbf{W-Rec.} &
\textbf{Macro-F1}
\tabularnewline
\hline

Joint Training & $52.0 \pm 1.0$ & $0.0 \pm 0.0$ & $52.4 \pm 1.0$ & $54.2 \pm 1.0$ & $52.0 \pm 1.0$ & $52.4 \pm 1.0$ \tabularnewline

Fine-Tuning & $34.9 \pm 2.0$ & $17.0 \pm 3.0$ & $35.1 \pm 2.0$ & $40.8 \pm 4.0$ & $34.9 \pm 2.0$ & $35.1 \pm 2.0$ \tabularnewline

EWC & $31.8 \pm 4.0$ & $20.2 \pm 4.0$ & $31.9 \pm 4.0$ & $34.5 \pm 4.0$ & $31.8 \pm 4.0$ & $31.9 \pm 4.0$ \tabularnewline

LwF & $27.2 \pm 4.0$ & $24.8 \pm 5.0$ & $26.3 \pm 3.0$ & $36.9 \pm 1.0$ & $27.2 \pm 4.0$ & $26.3 \pm 3.0$ \tabularnewline

ER & $38.4 \pm 2.0$ & $13.6 \pm 3.0$ &
$38.1 \pm 2.0$ & $39.7 \pm 2.0$ &
$38.4 \pm 2.0$ & $38.1 \pm 2.0$ \\

DER++ & $39.6 \pm 2.0$ & $12.4 \pm 2.0$ &
$39.5 \pm 2.0$ & $40.8 \pm 2.0$ &
$39.6 \pm 2.0$ & $39.5 \pm 2.0$ \\

DPDMR & $46.8 \pm 1.0$ & $5.2 \pm 1.0$ & $46.9 \pm 1.0$ & $49.2 \pm 3.0$ & $46.8 \pm 1.0$ & $46.8 \pm 1.0$ \tabularnewline

ProDER & \textbf{$51.0 \pm 1.0$} & \textbf{$1.0 \pm 1.0$} & \textbf{$51.4 \pm 1.0$} & \textbf{$53.6 \pm 1.0$} & \textbf{$51.0 \pm 1.0$} & \textbf{$51.3 \pm 1.0$} \tabularnewline

\hline
\end{tabular}%
}

\end{table*}

\subsection{Evaluation of Scenario 4 -- Fault Zone Localization with New Zone Labels}

Scenario 4 changes the prediction target from fault type classification to fault zone classification. For this reason, we evaluate it separately from the previous scenarios. Instead of distinguishing between several fault types, the model now predicts one of four zone labels. This makes the task easier in terms of the number of classes, but it still tests whether each CL method can learn new zone labels over time without forgetting previously learned zones.

The results are reported in Table~\ref{tab:scenario4_results} and Figures~\ref{sc4_1} and~\ref{sc4_2}. 
Scenario 4 shows a clear difference between methods that rely only on sequential updates and those that preserve information from previous tasks. Joint Training reaches an accuracy of $89.6 \pm 1.0$. In contrast, Fine-Tuning drops to only $25.6 \pm 0.0$ accuracy and a macro-F1 score of $11.2 \pm 0.0$. EWC and LwF improve the accuracy to $29.1 \pm 2.0$ and $33.4 \pm 1.0$, respectively, but their low macro-F1 scores indicate that regularization alone is still not sufficient to preserve balanced performance across the zone classes.

Replay-based methods perform considerably better. ER reaches an accuracy of $84.0 \pm 1.0$ and a macro-F1 score of $84.1 \pm 1.0$. DER++ and DPDMR obtain similar accuracies of $82.6 \pm 1.0$ and $82.6 \pm 3.0$, respectively. In this scenario, the additional mechanisms used by these methods do not provide an improvement over standard experience replay.

ProDER achieves the strongest CL result, reaching an accuracy of $91.4 \pm 1.0$ and a macro-F1 score of $91.4 \pm 1.0$. Its average performance is even slightly higher than Joint Training, resulting in a gap of $-1.8 \pm 1.0$. This should not be interpreted as ProDER being generally better than Joint Training, but it shows that the combination of replay, knowledge preservation, and prototype-based regularization is particularly effective for this task sequence. Overall, Scenario 4 shows that ProDER is not limited to fault type classification and can also maintain strong performance when the prediction target changes to fault zones.

\begin{figure}[H]
  \centering
  \begin{subfigure}[b]{0.49\textwidth}
    \includegraphics[width=\textwidth]{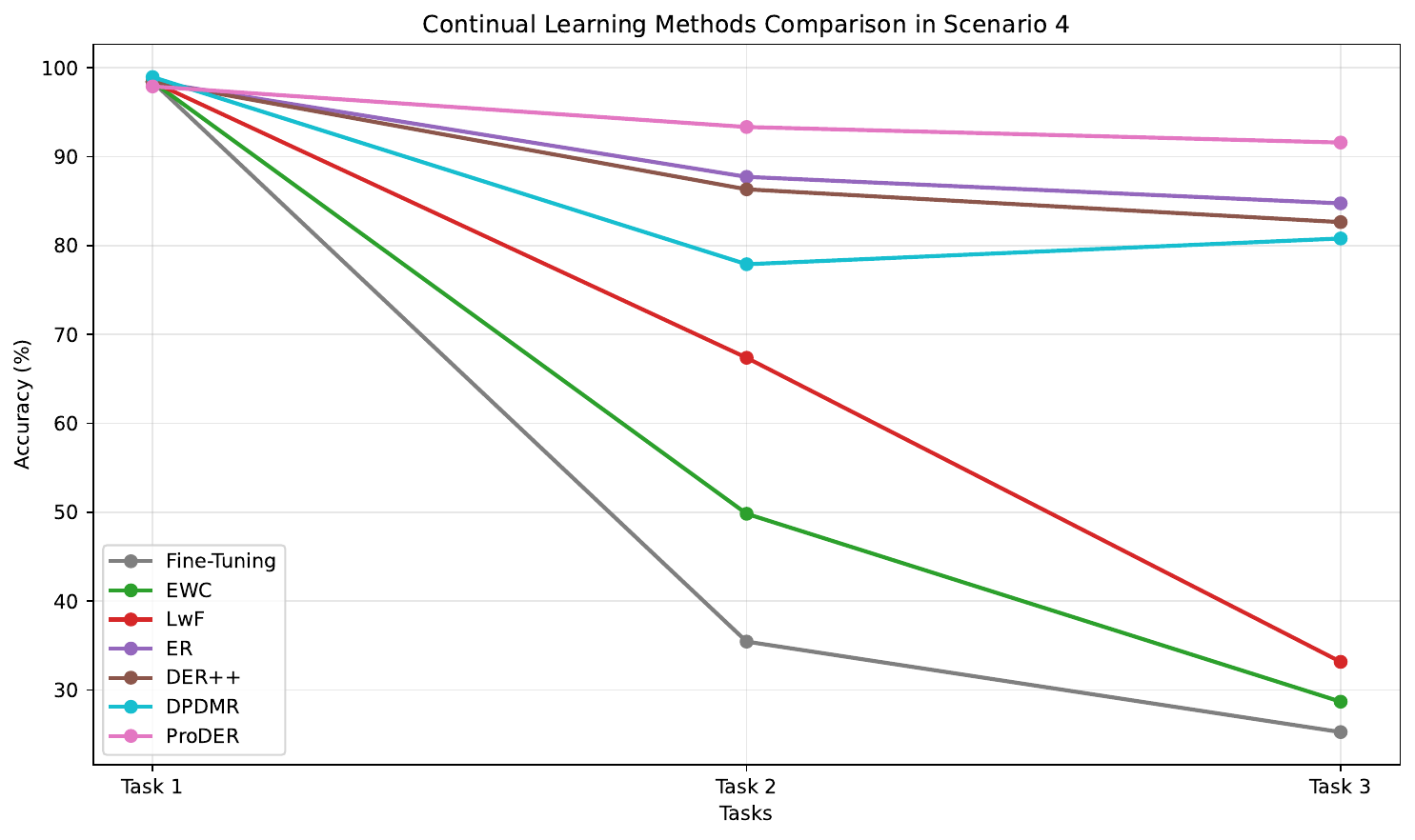}
    \caption{Test accuracy of each method and baselines}
    \label{sc4_1}
  \end{subfigure}
  \hfill
  \begin{subfigure}[b]{0.49\textwidth}
    \includegraphics[width=\textwidth]{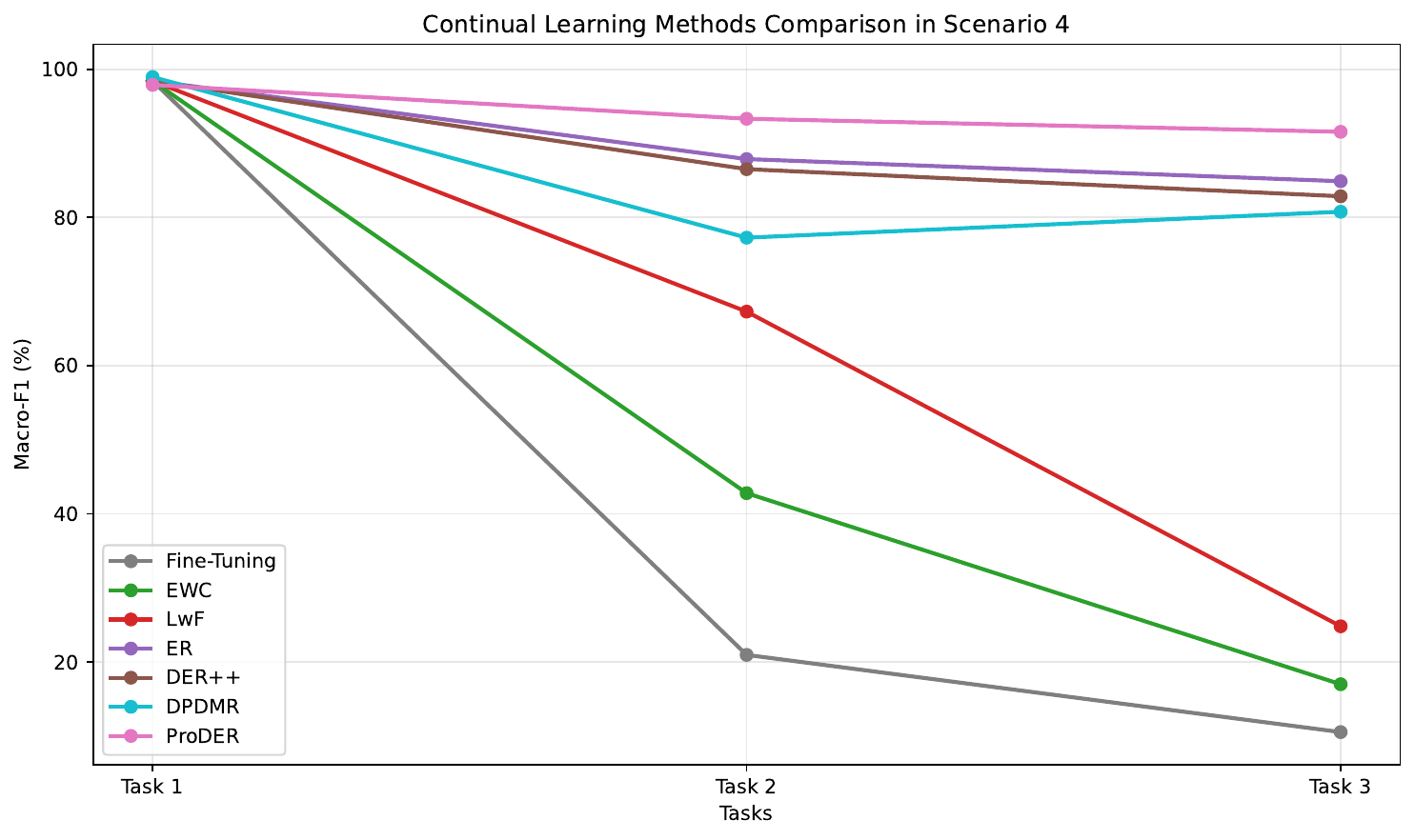}
    \caption{Test Macro-F1 of each method and baselines.}
    \label{sc4_2}
  \end{subfigure}
  \caption{Task-wise accuracy and Macro-F1 for one fixed seed in Scenario 4.}
  \label{fig:res_1}
\end{figure}

\begin{table*}[t]
\centering
\caption{Evaluation of CL methods in Scenario 4 using  accuracy, gap, and class-balanced classification metrics (Mean over 5 seeds). }
\label{tab:scenario4_results}

\scriptsize
\renewcommand{\arraystretch}{1.15}
\setlength{\tabcolsep}{3pt}

\resizebox{\textwidth}{!}{%
\begin{tabular}{lcccccc}
\hline
\textbf{Method} &
\textbf{ACC} &
\textbf{Gap} &
\textbf{W-F1} &
\textbf{W-Prec.} &
\textbf{W-Rec.} &
\textbf{Macro-F1}
\tabularnewline
\hline

Joint Training & $89.6 \pm 1.0$ & $0.0 \pm 0.0$ & $89.6 \pm 1.0$ & $89.7 \pm 1.0$ & $89.6 \pm 1.0$ & $89.6 \pm 1.0$ \tabularnewline

Fine-Tuning & $25.6 \pm 0.0$ & $64.0 \pm 1.0$ & $11.2 \pm 0.0$ & $31.3 \pm 0.0$ & $25.6 \pm 0.0$ & $11.2 \pm 0.0$ \tabularnewline

EWC & $29.1 \pm 2.0$ & $60.5 \pm 2.0$ & $17.7 \pm 3.0$ & $55.9 \pm 2.0$ & $29.1 \pm 2.0$ & $17.7 \pm 3.0$ \tabularnewline

LwF & $33.4 \pm 1.0$ & $56.2 \pm 1.0$ & $24.4 \pm 1.0$ & $65.6 \pm 13.0$ & $33.4 \pm 1.0$ & $24.4 \pm 1.0$ \tabularnewline

ER & $84.0 \pm 1.0$ & $5.6 \pm 1.0$ & $84.1 \pm 1.0$ & $85.4 \pm 1.0$ & $84.0 \pm 1.0$ & $84.1 \pm 1.0$ \tabularnewline

DER++ & $82.6 \pm 1.0$ & $6.9 \pm 1.0$ & $82.6 \pm 1.0$ & $83.2 \pm 1.0$ & $82.6 \pm 1.0$ & $82.6 \pm 1.0$ \tabularnewline

DPDMR & $82.6 \pm 3.0$ & $7.0 \pm 4.0$ & $82.7 \pm 3.0$ & $85.9 \pm 2.0$ & $82.6 \pm 3.0$ & $82.7 \pm 3.0$ \tabularnewline

ProDER & \textbf{$91.4 \pm 1.0$} & \textbf{$-1.8 \pm 1.0$} & \textbf{$91.4 \pm 1.0$} & \textbf{$91.9 \pm 1.0$} & \textbf{$91.4 \pm 1.0$} & \textbf{$91.4 \pm 1.0$} \tabularnewline

\hline
\end{tabular}%
}

\end{table*}

\subsection{Ablation Study}

The ablation study examines the contribution of the main components introduced in ProDER while keeping the training setup, replay-buffer budget, and shared hyperparameters unchanged. Starting from DER++, we first replace the original distillation formulation with KL-based distillation and use a joint cross-entropy loss over current and replayed samples. We then evaluate the semantic losses with random reservoir memory, followed by variants that use prototype-aware memory together with either the attraction or repulsion loss. Finally, we compare these variants with the complete ProDER framework, which combines KL-based distillation, joint classification, attraction and repulsion losses, and prototype-aware memory selection. The results are reported in Table~\ref{tab:ablation_proder}.

The results show that changing the DER++ objective alone has only a limited and scenario-dependent effect. The KL + Joint CE variant improves over DER++ in Scenarios 1, 2, and 4, but performs worse in Scenario 3. Larger gains appear when the semantic losses are introduced. Using the semantic losses with random reservoir sampling increases the accuracy from 40.1 to 44.8 in Scenario~1, from 39.9 to 42.5 in Scenario~2, and from 82.6 to 90.8 in Scenario~4. This indicates that the representation-level constraints contribute substantially beyond the changes to the replay objective.

The comparison between ER and DER++ also provides an indication of the contribution of output-level knowledge preservation. Since both methods use the same replay budget, while DER++ additionally preserves the stored outputs of previous samples, DER++ improves over ER by 7.3, 7.2, and 1.2 percentage points in Scenarios 1–3, respectively. In Scenario 4, however, DER++ is 1.4 points below ER. This suggests that distillation is particularly useful in the fault-type classification scenarios with repeated class updates, although its benefit is not uniform across all task sequences.

The attraction loss appears to be particularly important when considered on its own. Combined with prototype-aware memory, it reaches accuracies of 45.5, 43.5, 51.5, and 89.8 across the four scenarios. In contrast, the repulsion-only variant remains close to the DER++ baseline in several cases and performs below it in Scenario~2. The attraction and repulsion results also clarify why the two losses are more useful together than separately. Attraction first makes the representation of each class more compact around its prototype. Repulsion can then increase the separation between these compact groups by pushing their prototypes apart. If the samples of a class remain widely dispersed, however, separating the prototypes does not necessarily increase the separation between the samples themselves. This helps explain why repulsion alone provides little benefit, and even reduces performance in Scenario 2, while it becomes more useful when combined with attraction.

The contribution of the memory-selection strategy can be isolated by comparing Full ProDER with Semantic Losses + Reservoir, since these two variants use the same semantic losses and differ only in the replay selection rule. Prototype-aware selection improves accuracy by 1.3, 1.7, 3.8, and 0.6 percentage points in Scenarios 1–4, respectively. The largest improvement occurs in Scenario 3, where each fault class is observed across several grid zones. In this setting, samples belonging to the same class contain additional zone-dependent variation. Retaining both samples close to the prototype and samples farther from it can therefore provide broader coverage of this variation than random reservoir sampling, which is consistent with the larger gain observed in this scenario.

We do not include a separate memory-only variant in which prototype-aware selection is applied directly to the unregularized DER++ representation. The selection rule relies on distances in the learned embedding space, and these distances become more meaningful once the semantic losses have encouraged compact and separated class representations. Applying the same selection rule before structuring the embedding would therefore mix the quality of the representation with the quality of the selection strategy. We instead isolate memory selection by keeping the semantic losses fixed and changing only the replay selection rule.

Full ProDER achieves the highest mean accuracy in Scenarios~1, 2, and 4, reaching 46.1, 44.2, and 91.4, respectively. In Scenario~3, the attraction-only variant obtains a slightly higher mean accuracy than Full ProDER (51.5 versus 51.0), although Full ProDER has lower variability across seeds ($\pm 1.0$ compared with $\pm 2.0$). The ablation results show that the improvement of ProDER does not come from a single modification. The semantic losses provide most of the gain over DER++, while prototype-aware memory and the combined attraction--repulsion formulation generally provide additional improvements or more stable performance depending on the task sequence.

\begin{table*}[t]
\centering
\caption{Ablation study of the main components of ProDER. Results are reported as mean $\pm$ standard deviation over five seeds.}
\label{tab:ablation_proder}

\scriptsize
\renewcommand{\arraystretch}{1.15}
\setlength{\tabcolsep}{5pt}

\begin{tabular}{lcccc}
\hline
\textbf{Method} &
\textbf{Scenario 1} &
\textbf{Scenario 2} &
\textbf{Scenario 3} &
\textbf{Scenario 4}
\tabularnewline
\hline

DER++
& $40.1 \pm 1.0$
& $39.9 \pm 2.0$
& $39.6 \pm 2.0$
& $82.6 \pm 1.0$
\tabularnewline

DER++ + KL + Joint CE
& $41.0 \pm 2.0$
& $40.4 \pm 2.0$
& $37.8 \pm 2.0$
& $84.9 \pm 1.0$
\tabularnewline

Semantic Losses + Reservoir
& $44.8 \pm 2.0$
& $42.5 \pm 1.0$
& $47.2 \pm 3.0$
& $90.8 \pm 2.0$
\tabularnewline

Attraction + Prototype-Aware Memory
& $45.5 \pm 2.0$
& $43.5 \pm 3.0$
& $51.5 \pm 2.0$
& $89.8 \pm 2.0$
\tabularnewline

Repulsion + Prototype-Aware Memory
& $40.5 \pm 3.0$
& $34.9 \pm 3.0$
& $39.7 \pm 2.0$
& $82.6 \pm 2.0$
\tabularnewline

Full ProDER
& $46.1 \pm 2.0$
& $44.2 \pm 2.0$
& $51.0 \pm 1.0$
& $91.4 \pm 1.0$
\tabularnewline

\hline
\end{tabular}

\end{table*}

\subsection{Computational Runtime}
All runtime experiments were conducted on the Kaggle computing platform using an Intel(R) Xeon(R) CPU running at 2.00~GHz and an NVIDIA Tesla T4 GPU. The environment used Python 3.11.11 with PyTorch and CUDA 12.4. All evaluated methods were executed under the same computational setting to ensure a consistent runtime comparison.

Figure~\ref{fig:runtime_comparison} compares the total training time of the evaluated methods over the complete task sequence in each scenario. As expected, Fine-Tuning, ER, and LwF have the lowest computational cost, since they introduce either no additional optimization step or only a relatively lightweight replay or distillation mechanism. EWC requires a higher training time, particularly in Scenario~3, due to the additional computation required for estimating and applying the Fisher-based regularization.

Among the replay-based methods, DER++ remains relatively inexpensive, with total runtimes between 76.33 and 97.88 seconds across the four scenarios. ProDER introduces a higher training cost, ranging from 118.80 seconds in Scenario~4 to 156.62 seconds in Scenario~3. This increase is expected because, in addition to replay and logit distillation, ProDER computes feature representations and prototype-based attraction and repulsion terms during training. Nevertheless, its computational overhead remains moderate relative to DPDMR. DPDMR requires 218.41, 233.28, 322.60, and 179.23 seconds in Scenarios 1–4 whereas ProDER requires 150.11, 149.35, 156.62, and 118.80 seconds. The difference is particularly pronounced in Scenario~3, where DPDMR takes more than twice the training time of ProDER.

These results show that the improved CL performance of ProDER comes with additional computation compared with simpler replay methods such as ER and DER++. However, this additional cost remains bounded and is considerably lower than that of DPDMR in all four scenarios. This provides a practical trade-off between predictive performance, memory usage, and training cost without requiring repeated training over the full history of previously observed data.

\begin{figure*}[t]
    \centering
    \includegraphics[width=\textwidth]{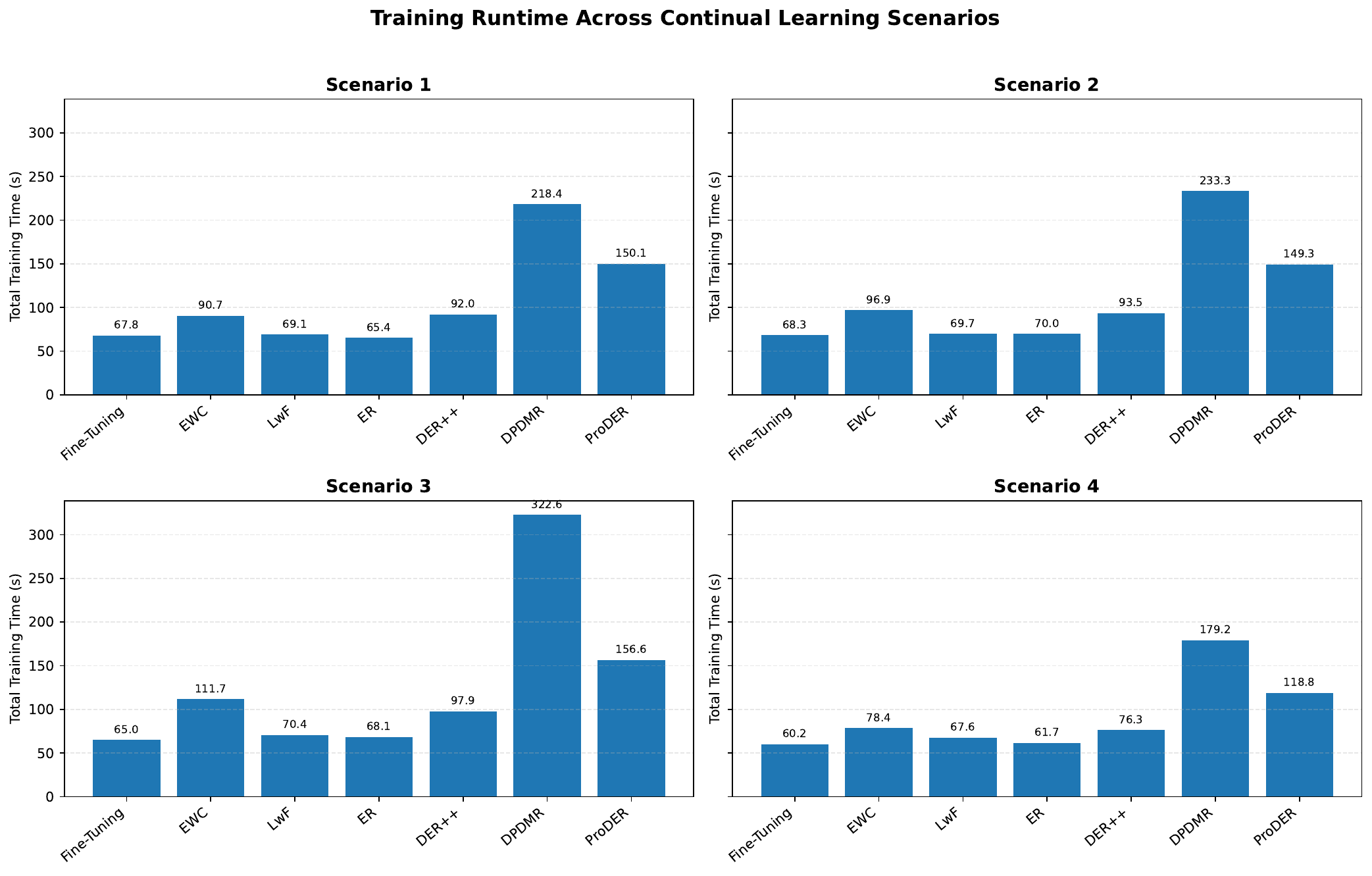}
    \caption{Total training runtime of the evaluated continual-learning methods across the four experimental scenarios. The reported values correspond to the cumulative training time over all tasks within each scenario.}
    \label{fig:runtime_comparison}
\end{figure*}

\subsection{Robustness under tighter replay-buffer budgets}

We further evaluate ER, DER++, DPDMR, and ProDER under a smaller replay-memory budget to examine how sensitive the methods are to limited access to past data. The main experiments use a buffer of 363 samples, which is reduced to 242 samples (15.8\% of training data) in this analysis. For Scenario 3, the corresponding buffer sizes are 353 and 236 samples (15.8\% of training data). The results are reported in Table~\ref{Tab:tab_results1}.

Reducing the buffer generally lowers performance, although the effect depends on the scenario and the replay strategy. The largest drops appear in Scenarios~1 and~2. For example, ProDER decreases from 46.1 to 40.6 in Scenario~1 and from 44.2 to 36.3 in Scenario~2. ER, DER++, and DPDMR show a similar trend in these two scenarios. In Scenario 3, however, increasing the buffer provides little or no improvement: ER and DER++ remain almost unchanged, while ProDER reaches the same mean accuracy of 51.0 with both buffer sizes. A similar effect is observed for DPDMR in Scenario 4, where both memory budgets result in an accuracy of 82.6 because the prototype-guided selection produces an effective replay set that already fits within the smaller buffer.

ProDER remains the strongest method under the smaller memory budget, achieving the highest mean accuracy in all four scenarios. Its advantage is particularly clear in Scenarios 2--4. Together with the ablation results in Section 6.5, these results show that the full ProDER framework remains effective when the available replay memory is reduced. The benefit of additional memory, however, depends on the task sequence and the replay method. ProDER retains samples according to their relation to the learned class prototypes while the attraction and repulsion losses help maintain a structured embedding space. The results therefore indicate that ProDER remains effective when the replay budget is reduced, although the benefit of additional memory naturally depends on the task sequence and how much of the available buffer is actually used.

\begin{table*}[htbp]
\centering
\caption{Performance of replay-based CL methods under two replay buffer sizes. Results are reported as mean $\pm$ standard deviation over five seeds.}
\label{Tab:tab_results1}

\scriptsize
\renewcommand{\arraystretch}{1.15}
\setlength{\tabcolsep}{8pt}

\begin{tabular}{lcccc}
\hline
\textbf{Method} &
\textbf{Scenario 1} &
\textbf{Scenario 2} &
\textbf{Scenario 3}\textsuperscript{3} &
\textbf{Scenario 4}
\tabularnewline

&
\textbf{ACC $\uparrow$} &
\textbf{ACC $\uparrow$} &
\textbf{ACC $\uparrow$} &
\textbf{ACC $\uparrow$}
\tabularnewline
\hline

ER\textsuperscript{1}
& $31.0 \pm 1.0$
& $26.2 \pm 0.0$
& $38.0 \pm 1.0$
& $81.3 \pm 2.0$
\tabularnewline

ER\textsuperscript{2}
& $32.8 \pm 1.0$
& $32.7 \pm 3.0$
& $38.4 \pm 2.0$
& $84.0 \pm 1.0$
\tabularnewline
\hline

DER++\textsuperscript{1}
& $38.6 \pm 1.0$
& $34.8 \pm 2.0$
& $40.7 \pm 1.0$
& $82.2 \pm 1.0$
\tabularnewline

DER++\textsuperscript{2}
& $40.1 \pm 1.0$
& $39.9 \pm 2.0$
& $39.6 \pm 2.0$
& $82.6 \pm 1.0$
\tabularnewline
\hline

DPDMR\textsuperscript{1}
& $37.2 \pm 2.0$
& $29.8 \pm 2.0$
& $47.0 \pm 2.0$
& $82.6 \pm 3.0$
\tabularnewline

DPDMR\textsuperscript{2}
& $40.9 \pm 2.0$
& $35.9 \pm 3.0$
& $46.8 \pm 1.0$
& $82.6 \pm 3.0$
\tabularnewline
\hline

ProDER (ours)\textsuperscript{1}
& \textbf{$40.6 \pm 2.0$}
& \textbf{$36.3 \pm 2.0$}
& \textbf{$51.0 \pm 2.0$}
& \textbf{$84.9 \pm 2.0$}
\tabularnewline

ProDER (ours)\textsuperscript{2}
& \textbf{$46.1 \pm 2.0$}
& \textbf{$44.2 \pm 2.0$}
& \textbf{$51.0 \pm 1.0$}
& \textbf{$91.4 \pm 1.0$}
\tabularnewline

\hline
\end{tabular}

\vspace{4pt}

\footnotesize
\textsuperscript{1} Replay buffer size = 242 samples. \quad
\textsuperscript{2} Replay buffer size = 363 samples. \quad
\textsuperscript{3} For Scenario 3, the corresponding replay buffer sizes are 236 and 353 samples, respectively.

\end{table*}

\subsection{Discussion}
In this section, we analyze the experimental results and their broader implications. Table~\ref{Tab:tab_results} summarizes all the results from different scenarios. Averaged over the four scenarios, ProDER reaches an accuracy of $58.2$ and a gap of $3.2$ to Joint Training, compared with $51.6$ and $9.8$ for DPDMR, the strongest competing replay method, and $50.6$ and $10.8$ for DER++. Since the accuracies in Scenario 4 are considerably higher than in the other scenarios, the average is mainly useful for comparing the methods with each other.

\begin{table*}[tbh]
\caption{Summary of the performance of the evaluated CL techniques across the four scenarios. ACC is the mean accuracy, and Joint Training is included as a reference. Avg.\ is the mean over the four scenarios. The best mean accuracy among the CL methods in each column is highlighted in \textbf{bold}.}
\label{Tab:tab_results}

\begin{adjustbox}{center, max width=\textwidth}
\begin{tabular}{|c|c|c|c|c|c|}
\hline

& \textbf{Scenario 1}
& \textbf{Scenario 2}
& \textbf{Scenario 3}
& \textbf{Scenario 4}
& \textbf{Avg.}
\tabularnewline

\hline

& \textbf{ACC $\uparrow$}
& \textbf{ACC $\uparrow$}
& \textbf{ACC $\uparrow$}
& \textbf{ACC $\uparrow$}
& \textbf{ACC $\uparrow$}
\tabularnewline

\hline

\textbf{Joint Training}
& 52.0 & 52.0 & 52.0 & 89.6 & 61.4
\tabularnewline



\hline

\textbf{Fine-Tuning}
& 15.4 & 9.3 & 34.9 & 25.6 & 21.3
\tabularnewline

\hline

\textbf{EWC}
& 14.9 & 9.9 & 31.8 & 29.1 & 21.4
\tabularnewline

\hline

\textbf{LwF}
& 16.5 & 9.2 & 27.2 & 33.4 & 21.6
\tabularnewline

\hline

\textbf{ER}
& 32.8 & 32.7 & 38.4 & 84.0 & 47.0
\tabularnewline

\hline

\textbf{DER++}
& 40.1 & 39.9 & 39.6 & 82.6 & 50.6
\tabularnewline

\hline

\textbf{DPDMR}
& 40.9 & 35.9 & 46.8 & 82.6 & 51.6
\tabularnewline

\hline

\textbf{ProDER (ours)}
& \textbf{46.1} & \textbf{44.2} & \textbf{51.0} & \textbf{91.4} & \textbf{58.2}
\tabularnewline

\hline

\end{tabular}
\end{adjustbox}
\end{table*}

When examining the performance of regularization-based methods like EWC and LwF, it becomes clear that they struggle across most scenarios, showing only marginal improvements over Fine-Tuning (the lower bound). This underperformance can be attributed to their underlying mechanisms being ill-suited to the type of challenges presented in our benchmarks, particularly the frequent introduction of new classes.

In contrast, ProDER, DER++, and ER use a fixed replay capacity of 363 samples for scenario 1, 2 and 4 , also 353 samples for scenario 3, all corresponding to 23.7\% of the training-set size. ProDER additionally maintains a class-balanced allocation, corresponding to up to 33 exemplars per class once all 11 classes have been observed. This makes them much more realistic and scalable for deployment in resource-constrained environments. Across all scenarios, ProDER consistently outperforms all the tested CL approaches.

The relatively modest Joint Training performance for fault type classification should also be considered in the context of the underlying classification task. Unlike fault zone localization, which contains four output classes, fault type classification requires discrimination among 11 different fault categories. Using the same network architecture, Joint Training reaches 89.6\% accuracy on the fault zone localization task, indicating that the lower ceiling is specific to the more difficult fault type classification problem rather than to the network itself. Moreover, each fault type is represented through a limited set of simulated operating conditions, including 22 resistance values. Consequently, the fault type task remains challenging even when all classes are jointly available during training. The Joint Training result should therefore be interpreted as an empirical reference for the considered dataset and model configuration and not as an estimate of the maximum achievable accuracy for fault diagnosis in general.
\subsection{Limitations and Future Work}
\label{sec:limitations}

All experiments use simulated data from the IEEE 13-node test feeder~\cite{mainpaper}, the only public dataset we are aware of that annotates both fault type and fault zone for each event. 
Real deployments may include measurement noise, missing or corrupted sensor readings, communication delays, changing loading conditions, variations in distributed energy resources, and protection-device behavior that are difficult to represent completely in simulation.
In addition, a single feeder with four zones cannot represent the variety of topologies, voltage levels, and penetration of distributed generation found in real distribution networks. Consequently, the absolute accuracies reported here should not be taken as estimates of field performance; what our results support is the \textit{relative} comparison among CL strategies under controlled and identical conditions.
Similarly, we did not evaluate robustness to measurement noise, delayed or partial labels, or topology reconfigurations, since these factors are not present in the dataset.

Second, the experiments consider one feeder configuration with four fault zones and eleven fault types. The effectiveness of the learned representations and prototype-based replay may therefore vary for larger networks, different topologies, or systems with substantially different operating conditions. The experiments also assume a predefined sequence of continual-learning tasks, whereas in practice distribution changes may occur gradually and without explicit task boundaries.

These limitations suggest several directions for future work. A primary next step is to evaluate ProDER using real utility measurements, hardware-in-the-loop platforms, or high-fidelity digital-twin environments in which realistic sensor noise, communication effects, and changing operating conditions can be introduced. Evaluation across multiple feeder topologies and different levels of renewable and distributed-energy penetration would also help assess the generalizability of the framework. Another direction is to extend ProDER toward task-free continual learning, where changes in the operating distribution are detected automatically rather than being provided through predefined task boundaries. Finally, future work should investigate robustness to noisy, missing, and imbalanced observations and explore adaptive replay-memory strategies that dynamically adjust the memory budget according to the severity of distribution change and the amount of forgetting observed during operation.

\section{Conclusion}
\label{Con}

This paper studied continual learning for fault diagnosis in smart grids, where new fault types appear and monitoring is extended to new grid zones over time. We formulated fault type classification and fault zone localization as continual learning problems and defined four evaluation scenarios on the IEEE 13-node test feeder, three class-incremental and one domain-incremental. To mitigate catastrophic forgetting under a fixed replay budget, we proposed Prototype-based Dark Experience Replay (ProDER), which extends DER++ with prototype attraction and repulsion losses, temperature-scaled logit distillation, and a prototype-aware replay memory that retains both core and boundary samples of each class.

Our experiments lead to three main findings. First, regularization-based methods such as EWC and LwF improve only marginally over Fine-Tuning, whereas replay-based methods retain considerably more knowledge, which makes replay essential when new classes are introduced. Second, ProDER achieved the highest mean accuracy among the continual learning methods in every scenario ($46.1$, $44.2$, and $51.0$ for fault type classification, and $91.4$ for fault zone localization), outperforming the strongest competing method by $4.2$ to $7.4$ percentage points. Its gap to Joint Training was $5.8$, $7.8$, and $1.0$ points in the fault-type scenarios.

Third, the ablation study showed that the improvement does not stem from a single component: the prototype-based losses provide most of the gain over DER++, the attraction loss being the most consistently beneficial, and the full combination was the best configuration in three of the four scenarios and the most stable in the fourth. ProDER also remained the strongest method when the replay buffer was reduced by one third, from $363$ to $242$ samples.

These gains come at a moderate and bounded cost. ProDER is more expensive to train than ER and DER++, since it computes feature-level regularization terms and prototype-aware memory updates, but it was faster than DPDMR in all four scenarios and it operates with a fixed replay buffer.
Overall, the results suggest that combining replay with structured feature-space regularization is a practical, memory-bounded way to keep fault diagnosis models up to date as the grid evolves. Since all experiments rely on simulated data from a single feeder, and even Joint Training reaches only about $52$ for the eleven-class fault-type task, the reported accuracies support a relative comparison among continual learning strategies rather than an estimate of field performance. Validating ProDER on real measurements, across different topologies, and in task-free settings, as discussed above, is the natural next step.

\section*{Declarations}

\subsection*{Funding}
No funding was received for conducting this study.

\subsection*{Competing interests}
The authors declare that they have no competing interests.

\subsection*{Ethics approval and consent to participate}
Not applicable.

\subsection*{Consent for publication}
Not applicable.

\subsection*{Data availability}
The data supporting the findings of this study are public.

\subsection*{Materials availability}
Not applicable.

\subsection*{Code availability}
The code used in this study is provided.

\subsection*{Author contribution}
All authors contributed to the conception and design of the study. Material preparation, data processing, implementation, experiments, and analysis were performed by the authors. The first draft of the manuscript was written by the authors, and all authors reviewed and approved the final manuscript.

\bibliography{sn-bibliography.bib}

\end{document}